\documentclass{article}

\usepackage[final]{neurips_2021}

\usepackage[utf8]{inputenc} 
\usepackage[T1]{fontenc}    
\usepackage{hyperref}       
\usepackage{url}            
\usepackage{booktabs}       
\usepackage{amsfonts}       
\usepackage{nicefrac}       
\usepackage{microtype}      
\usepackage{xcolor}         
\usepackage{wrapfig}

\usepackage{mystyle}
\usepackage{algorithm}
\usepackage[noend]{algorithmic}

\usepackage{subcaption}
\usepackage{multirow}

\newcommand\blfootnote[1]{%
  \begingroup
  \renewcommand\thefootnote{}\footnote{#1}%
  \addtocounter{footnote}{-1}%
  \endgroup
}

\title{Low-Rank Constraints for Fast Inference in Structured Models}

\author{%
  Justin T. Chiu\thanks{Equal contribution}\\
  Cornell University \\
  \texttt{jtc257@cornell.edu} \\
  \And
  Yuntian Deng\footnotemark[1] \\
  Harvard University \\
  \texttt{dengyuntian@seas.harvard.edu} \\
  \AND
  Alexander M. Rush \\
  Cornell University \\
  \texttt{arush@cornell.edu} \\
}

\begin{document}

\maketitle

\begin{abstract}
Structured distributions, i.e. distributions over combinatorial spaces, are commonly used to learn latent probabilistic representations from observed data. However, scaling these models is bottlenecked by the high computational and memory complexity with respect to the size of the latent representations. Common models such as Hidden Markov Models (HMMs) and Probabilistic Context-Free Grammars (PCFGs) require time and space quadratic and cubic in the number of hidden states respectively. This work demonstrates a simple approach to reduce the computational and memory complexity of a large class of structured models. We show that by viewing the central inference step as a matrix-vector product and using a low-rank constraint, we can trade off model expressivity and speed via the rank.  Experiments with neural parameterized structured models for language modeling, polyphonic music modeling, unsupervised grammar induction, and video modeling show that our approach matches the accuracy of standard models at large state spaces while providing practical speedups.
\end{abstract}

\section{Introduction}


When modeling complex sequential spaces, such as sentences, musical scores, or video frames,
a key choice is the internal structural representations of the model. A common choice in recent years is to use neural representations~\citep{nplm,mikolov-2011,gpt3,polyphonic,music-trans,video} to store a deterministic history. These models yield strong predictive accuracy but their deterministic, continuous forms provide little insight into the intermediate decisions of the model.\blfootnote{Code is available \href{https://github.com/justinchiu/low-rank-models}{\underline{here}}.}

Latent structured models provide an alternative approach where complex modeling decisions are broken down into a series of probabilistic steps. Structured models provide a principled framework for reasoning about the probabilistic dependencies between decisions and for computing posterior probabilities. 
The structure of the decision processes and the ability to answer queries through probabilistic inference afford interpretability and controllability that are lacking in neural models~\citep{pgm,levine2018reinforcement}.
 

Despite the benefits of structured models, the computational complexity of training scales asymptotically much worse than for neural models, as inference, and therefore training, requires marginalizing over all possible latent structures.
For standard general-purpose models like Hidden Markov Models (HMM) and Probabilistic Context-Free Grammars (PCFG), the runtime of inference scales quadratically and cubically in the number of states respectively, which limits the ability to reach a massive scale.
Promisingly, recent work has shown that in specific situations these models can be scaled, and that the increased scale results in commensurate improvements in accuracy -- without sacrificing the ability to perform exact inference~\citep{dedieu2019learning,chiu2020scaling,yang2021pcfgs}.

In this work, we propose an approach for improving the runtime of a large class of structured latent models by introducing a low-rank constraint. 
We target the family of models where inference can be formulated through a
labeled directed hypergraph, which describes a broad class
of dynamic-programming based inference~\citep{klein2004parsing,huang2005better,zhou2006learning,javidian2020hypergraph,chiang2020factor}. We show how under low-rank constraints these models allow for more efficient inference. 
Imposing a low-rank constraint allows for a key step of inference to be rewritten as a fast matrix-vector product.
This approach is also inspired by recent advances in computationally efficient neural attention  attention~\citep{katharopoulos2020lineartransformer,peng2021rfa,choromanski2020performer},
a significantly different task and formulation, that rewrites matrix-vector products as fast low-rank products using approximate kernel techniques.

We evaluate this approach by learning low-rank structured models for the tasks of language modeling, polyphonic music modeling, unsupervised grammar induction, and video modeling. For these tasks we use a variety of models including HMMs, PCFGs, and Hidden Semi-Markov Models (HSMMs).
As the application of low-rank constraints is nontrivial in high-dimensional structured models due to reduced expressivity, we demonstrate effective techniques for overcoming several practical challenges of low-rank parameterizations.
We find that our approach achieves very similar results to unconstrained models at large state sizes, while the decomposition allows us to greatly increase the speed of inference.
Results on HMMs show that we can scale to more than 16,000 states; results on PCFGs achieve a significant perplexity reduction from much larger state spaces compared to past work \citep{kim2019cpcfg}; and results on HSMMs show that our formulation enables scaling to much larger state spaces for continuous emissions~\citep{fried2020learning}.

\section{Background: Latent Structure and Hypergraphs}

We consider the problem of modeling a sequence of observations $p(x)= p(x_1, \dots, x_T)$.
These observations can range in complexity from the words in a sentence to a series of co-occurring musical notes,
or to features of video frames, and 
may be discrete or continuous.
We assume these observations are generated by an unobserved (latent) structured representation $z$, and therefore model the joint
$p(x, z)$.
The structure may be sequential or hierarchical, such as latent trees,
and the set of structures $\mcZ$ is combinatorial, i.e. exponential in size with respect to the input sentence. 
In order to train these models on observations, we must optimize the evidence
$p(x) = \sum_z p(x,z)$ by marginalizing over $z$. Scaling this marginalization is the focus of this work. 

Hypergraphs are a graphical model formalism  for structured distributions that admit tractable
inference through dynamic programming~ \citep{klein2004parsing,huang2005better,zhou2006learning,javidian2020hypergraph,chiang2020factor}.\footnote{While the formalism is similar to undirected factor graphs,
it allows us to represent more complex distributions: notably
dependency structures with unknown topologies, such as latent trees.}
A labeled, directed, acyclic hypergraph consists of a set of nodes
$\cal V$, a set of hyperedges $\cal E$, and a designated root node $S \in {\cal V}$. 
Each node $v \in \mcV$ has a collection of labels ${\cal L}_v$.
Each hyperedge $e\in\mcE$ has a head node $u$ and tuple of tail nodes, $v = (v_1, \ldots, v_{|e|})$, where $|e|$ is the number of tail nodes.
For simplicity, we will assume \textit{at most} 2 tail nodes $v_1, v_2$, and unless noted, a fixed label set $\mcL$ throughout.
Each hyperedge $e$ is associated with a score matrix $\Psi_e\in\R^{\mcL\times \mcL^{|e|}}$ 
with a score for all head and tail labels.\footnote{This formalism can represent inference in both locally and globally normalized models, although we focus on local normalization in this work.}
We use the notation $[\Psi_e]_{z_u,(z_1, z_2)}$ to indicate the score for head label $z_u$ and tail labels $z_1$ and $z_2$.
Finally, we assume we have a topological ordering over the edges.

A hypergraph is used to aggregate scores bottom-up through a dynamic programming (belief propagation) algorithm.
Algorithm~\ref{lab:hyp-matrix}~(left) shows the algorithm. It works by filling in a table vector
$\alpha_v \in \R^{\mcL}$ for each node $v$ in order, and is initialized to 1 at the leaf nodes.\footnote{
The accumulation of scores is denoted by $\stackrel{+}{\gets}$.
Multiple hyperedges can have the same head node, whose scores must be added together.}
It returns the sum over latent structures, $p(x)$. Counting
loops, the worst-case runtime complexity is
$O(|{\cal E}|\times L^{|e^*|+1})$ where $L = |{\cal L}|$ is the size of the
label set and $|e^*|$ the max hyperedge tail size. Algorithm~\ref{lab:hyp-matrix}~(right) shows the 
same algorithm in matrix form by introducing joined tail vectors $\beta_v\in\R^{\mcL^{|e|}}$ for each group of nodes $v$.  Letting $z_v = (z_1,z_2)$, the joined tail vector contains entries $[\beta]_{z_v} = [\alpha_{v_1}]_{z_1}[\alpha_{v_2}]_{z_2}$.

\begin{algorithm}[t]
\begin{minipage}[t]{0.50\textwidth}
\begin{algorithmic} 
\STATE{[\textit{Scalar Form}]}
\FOR {$u \leftarrow v_1, v_2$ hyperedge $e$ topologically}
\FOR {$z_u \in {\cal L}_u$ }
\STATE $[\alpha_u]_{z_u} \stackrel{+}{\gets}  \sum_{z_1, z_2}  [\Psi_e]_{z_u,(z_1, z_2)}$
\STATE \hspace{7em} $\bm\cdot \ [\alpha_{v_1}]_{z_1} \  [\alpha_{v_2}]_{z_2}$
\ENDFOR
\ENDFOR
\STATE \textbf{return} $\sum_z [\alpha_S]_z$
\end{algorithmic}
\end{minipage}
\begin{minipage}[t]{0.45\textwidth}
\begin{algorithmic} 
\STATE{[\textit{Matrix Form}]}
\FOR {$u \leftarrow v$ hyperedge $e$ topologically}
\STATE $\alpha_u \stackrel{+}{\gets} \Psi_e\beta_v$
\ENDFOR
\STATE \textbf{return} $\alpha_S^\top \mathbf{1}$

\vspace{1.5em}
\end{algorithmic}
\end{minipage}
\caption{\label{lab:hyp-matrix} Hypergraph marginalization }
\end{algorithm}

To make this formalism more concrete, we show how hypergraphs can be used for inference in several structured generative models: hidden Markov models,  probabilistic context-free grammars, and hidden semi-Markov models.
Inference in these examples are instances of the hypergraph algorithm.

\textbf{Example: Hidden Markov Models (HMM)}
HMMs are discrete latent sequence models defined by the following generative process: first, a sequence of discrete latent states $z = (z_1, \ldots,z_T)$ with state size $L$ are sampled as a Markov chain. Then each state $z_t$ independently emits an observation $x_t$, i.e. 
\begin{equation}
\label{eqn:hmm}
    p(x,z) = \prod_{t=1}^T p(z_t \mid z_{t-1})\ p(x_t\mid z_t),
\end{equation}
where $p(z_t \mid z_{t-1})$ is the transition distribution,
 $p(x_t \mid z_t)$ the emission distribution, and $p(z_1 \mid z_0)$ is the initial distribution with distinguished start symbol $z_0$.

Given a sequence of observations $x=(x_1, \ldots, x_n)$ we can compute
$p(x) = \sum_zp(x,z)$ using a labeled directed hypergraph, with single-tailed edges, nodes corresponding to state positions, labels corresponding to states, and emissions probabilities incorporated into the scoring matrices $\Psi$.
There are $T$ scoring matrices, $\Psi_t\in\R^{\mcL\times\mcL}$, with entries $[\Psi_t]_{z_t,z_{t+1}} = p(z_{t+1},x_t \mid z_t)$ corresponding to transitions.\footnote{
The left-most scoring matrix for the HMM has entries $[\Psi_1]_{z_1,z_{2}} = p(z_{2},x_1 \mid z_1)p(z_1\mid z_0)$.
}
Algorithm~\ref{fig:marg-hmm-pcfg}~(left) shows the approach. This requires time $O(TL^2)$ and is identical to the backward algorithm for HMMs.\footnote{
In the case of HMMs, the table vectors $\alpha_t$ correspond to the backward algorithm's $\beta$ values.}

\begin{algorithm}[t]
\caption{Hypergraph marginalization for HMMs and PCFGs}
\label{fig:marg-hmm-pcfg}
\begin{minipage}[t]{0.45\textwidth}
\begin{algorithmic}
\STATE {[\textit{HMM - Backward}]}
\FOR {$t \leftarrow (t+1)$ in right-to-left order}
\FOR {$z_{t+1} \in \mcL$}
\STATE $[\beta_{t+1}]_{z_{t+1}} = [\alpha_{t+1}]_{z_{t+1}}$
\ENDFOR
\STATE $\alpha_t \stackrel{+}{\gets} \Psi_t \beta_{t+1}$
\ENDFOR
\STATE \textbf{return} $\alpha_0^\top \mathbf{1}$
\end{algorithmic}
\end{minipage}
\vspace{0pt}
\begin{minipage}[t]{0.50\textwidth}
\begin{algorithmic} 
\STATE {[\textit{PCFG - CKY}]}
\FOR {$(i,k) \leftarrow (i,j), (j,k)$ in span-size order}
\FOR {$z_1,z_2 \in \mcL_{i,j}\times\mcL_{j,k}$}
\STATE $[\beta_{i,j,k}]_{(z_1,z_2)} = [\alpha_{i,j}]_{z_1}[\alpha_{j,k}]_{z_2}$
\ENDFOR
\STATE $\alpha_{i,k} \stackrel{+}{\gets} \Psi\beta_{i,j,k}$
\ENDFOR
\STATE \textbf{return} $\alpha_{1,T}^\top \mathbf{1}$
\end{algorithmic}
\end{minipage}
\end{algorithm}


\textbf{Example: Context-Free Grammars (CFG)}
CFGs are a structured model defined by the 5-tuple 
$\mcG = (S,\mcN,\mcP,\mcX,\mcR)$, where $S$ is the distinguished start symbol, $\mcN$ is a set of nonterminals, $\mcP$ is a set of preterminals, $\mcX$ is the token types in the vocabulary, and $\mcR$ is a set of grammar rules.
Production rules for start, nonterminals, and preterminals take the following forms:\footnote{
We restrict our attention to grammars in Chomsky normal form.}
\begin{equation}
\label{eqn:pcfg-rules}
\begin{aligned}
S&\to A, & A\in\mcN; &&
A&\to B\ C, & B,C\in\mcN\cup\mcP; &&
D&\to x, & D \in\mcP,x\in\mcX.
\end{aligned}
\end{equation}
A probabilistic context-free grammar (PCFG) additionally has a probability measure on the set of rules.
To compute $p(x_1, \ldots, x_T)$ with a hypergraph, we create one node for each contiguous
 subspan $[i, k)$ in the sentence. Nodes with $i + 1 < k$ have a
nonterminal label set ${\cal L} = \mcN$. Nodes with $i + 1=k$ have a 
preterminal label set ${\cal L}_{i,i+1} = \mcP$.
The main scoring matrix is $\Psi\in\R^{\mcL\times \mcL^2}$, with entries $[\Psi]_{z_u,(z_1,z_2)} = p(z_1,z_2\mid z_u)$.\footnote{We have a separate matrix for terminal production on $x$ which we elide for 
simplicity.}
Algorithm~\ref{fig:marg-hmm-pcfg} (right) shows how for every hyperedge we join the scores from the two tail nodes in $\alpha_{i,j}$ and $\alpha_{j,k}$ into joined tail vector $\beta_{i,j,k}\in\R^{\mcL^2}$. 
As there are $O(T^3)$ hyperedges and the largest ${\cal L}$ is of size
$|\mcN|$, the runtime of the algorithm is $O(T^3 {|\mcN|}^3)$. This approach is identical to the CKY algorithm.

\textbf{Example: Hidden Semi-Markov Models (HSMM)}
HSMMs are extensions of HMMs that allow for generating a variable-length sequence of observations per state. 
It defines the following generate process: first, we sample a sequence of discrete latent states $z=(z_1, \cdots, z_K)$ with a first-order Markov model. We then use them to generate the length of observations per state. For our experiments we generate independent continuous emissions $x_t$ with a Gaussian distribution for $p(x_i\mid z_k)$. Full details of the inference procedure are given in Appendix~\ref{sec:hsmm}.



\section{Rank-Constrained Structured Models}
\label{sec:linearizing}

For these structured distributions, hypergraphs provide a general method for inference (and therefore training parameterized versions). However, the underlying algorithms scale poorly with the size of the label sets (quadratic for HMM and HSMM, cubic for CFG). This complexity makes it challenging to scale these models and train versions with very large numbers of states.

In this section, we consider an approach for improving the scalability
of these models by reducing the dependence of the computational complexity of inference on the label set size.
The main idea is to speed up the matrix-vector product step in inference by using a low-rank decomposition of 
the scoring matrix $\Psi$.
In the next section we show that this constraint can be easily  incorporated into parameterized versions of these models.

\subsection{Low-Rank Matrix-Vector Products}
The main bottleneck for inference speed is the matrix-vector product $\alpha_u \stackrel{+}{\gets} \Psi_e\beta_v$ that must be computed for every edge in the hypergraph.
As we saw in Algorithm~\ref{lab:hyp-matrix} (left), this step takes time $L^{|e|+1}$ to compute,
but it can be sped up by making structural assumptions
on $\Psi_e$. In particular, we focus on scoring matrices with low rank.

We note the following elementary property of matrix-vector products.
If the scoring matrix can be decomposed as the product
of two smaller matrices $\Psi_e = U_e V_e^\top$,
where $U_e \in \mathbb{R}^{\mcL \times N}$ and $V_e \in \mathbb{R}^{N\times \mcL^{|e|}}$,
then the matrix-vector products can be computed in time
$O(|{\cal E}|\times L^{|e|}\times N)$ as follows:

\begin{equation}
    \label{eqn:hypergraph-update-kernel-matrix}
    \Psi_e\beta_v =  \left(U_e V_e^\top\right) \beta_v =  U_e\left(V_e^\top\beta_v\right).
\end{equation}

This reordering of computation exchanges a factor of $L$ for a factor of $N$. When $N \ll L$, this method is both faster and more memory-efficient. 

We enforce the low-rank constraint by directly parameterizing the factors $U_e$ and $V_e$ for scoring matrices $\Psi_e$ that we would like to constrain. We treat both $U_e$ and $V_e$ as embedding matrices, where each row corresponds to an embedding of each value of $z_u$ and a joint embedding of $(z_1,z_2)$ respectively:
\begin{equation}
    [U_e]_{z_u,n} = c_{z_u}[\phi(f(z_u))]_n 
    \qquad
    [V_e]_{(z_1,z_2),n} = c_{z_1,z_2} [\phi(g(z_1,z_2))]_n,
\end{equation}
where $f$ and $g$ are embedding functions; $c_{z_u}$ and $c_{z_1,z_2}$ are constants (used to ensure proper normalization) or clamped potentials (such as conditional probabilities); and $\phi:\R^D\to\R^N_+$ is a function that ensures nonnegativity, necessary for valid probability mass functions. 
Algorithm~\ref{alg:low-rank-update} shows the role of the low-rank
matrix-vector product in marginalization.\footnote{
If the normalizing constants are given by $c_{z_u}$,
they can be computed from unnormalized $\tilde{U}_e,\tilde{V}_e$ as follows: $c_{z_u} = [\tilde{U}_e\tilde{V}_e^\top\mathbf{1}]_{z_u}$ in time $O(L^{|e|}N + LN)$, and similarly for $c_{z_1,z_2}$.
}

\begin{wrapfigure}{r}{0.55\textwidth}
\begin{minipage}{0.55\textwidth}
\begin{algorithm}[H]
\caption{\label{alg:low-rank-update} Low-rank marginalization}
\begin{algorithmic} 
\FOR {$u \leftarrow v_1, v_2$ hyperedge $e$ topologically}
\FOR {$n \in 1,\ldots,N$}
\STATE \hspace*{-0.2cm}$[\gamma]_n = \displaystyle \sum_{z_v} c_v\  [\phi(g(z_{1}, z_{2}))]_n \  [\beta_{v}]_{z_v}$
\hfill $\vartriangleright$ $O(L^{|e|})$
\ENDFOR
\STATE $\alpha_u \stackrel{+}{\gets} U_e\gamma $
    \hfill $\vartriangleright$ $O(LN)$
\ENDFOR
\STATE \textbf{return} $\alpha_S^\top\mathbf{1}$
\end{algorithmic} 
\end{algorithm}
\end{minipage}
\end{wrapfigure}

\subsection{Application to Structured Models}
As enforcing a low-rank factorization of every scoring matrix limits the expressivity of a model,
we explicitly target scoring matrices that are involved in computational bottlenecks.\footnote{For a discussion of the expressivity of low-rank models compared to models with fewer labels, see Appendix~\ref{sec:expressivity}.}
For these key scoring matrices, we directly parameterize the scoring matrix with a low-rank factorization, which we call a low-rank parameterization.
For other computations, we utilize a standard softmax parameterization and do not factorize the resulting scoring matrix.
We refer to this as a mixed parameterization.

\textbf{Hidden Markov Models}
Low-rank HMMs (LHMMs) use the following mixed parameterization, which specifically targets the state-state transition bottleneck by using a low-rank parameterization for the transition distribution, but a softmax parameterization for the emission distribution:
\begin{equation}
\begin{aligned}
p(z_t \mid z_{t-1}) &\propto \phi(\bu_{z_{t-1}})^\top \phi(\bv_{z_t}),&
p(x_t \mid z_t) &\propto \exp(\bu_{z_t}^\top \bv_{x_t}),
\end{aligned}
\end{equation}
where $\bu_{z_{t-1}} = f(z_{t-1})$ and $\bv_{z_t} = g(z_{t})$ are (possibly neural) embedding functions.
The parameterizations of the embedding functions $f,g:\mcL\to\R^D$, as well as the non-negative mapping $\phi:\R^D\to\R^N_+$ are detailed in Appendix~\ref{sec:mlp-param}.
When performing inference, we treat the emission probabilities $p(x_t\mid z_t)$ as constants,
and absorb them into $c_u$.

This allows inference to be run in time $O(TLN)$, where $T$ is the length of a sequence, $L$ the size of the label space, and $N$ the feature dimension.

\textbf{Hidden Semi-Markov Models}
For low-rank HSMM (LHSMM),  we similarly target  the transition distribution and keep the standard Gaussian emission distribution:
\begin{equation}
\label{eqn:hsmm}
\begin{aligned}
p(z_k \mid z_{k-1}) &\propto \bu_{z_{k-1}}^\top \bv_{z_k},&
p(x_t \mid z_k) &\propto K_{\text{Gauss}}(\bu_{z_k}, \mathbf{x}_t),
\end{aligned}
\end{equation}
where $\bu_{z_{k-1}} = \phi(f(z_{k-1}))$ and $\bv_{z_k} = \phi(g(z_k))$ are state embeddings, while $K_{\text{Gauss}}(\cdot, \cdot)$ is the Gaussian kernel used to model continuous $\mathbf{x}_t$. The full parameterization of the embeddings is given in Appendix \ref{sec:mlp-param}. The total inference complexity is $O(TLMN )$, where $M$ is the maximum length of the observation sequence under any state.

\textbf{Context-Free Grammars}
For PCFGs, the inference bottleneck is related to the transition from a nonterminal symbol to two nonterminal symbolss ($A\to B\ C$), and we specifically parameterize it using a low-rank parameterization:
\begin{equation}
\label{eqn:pcfg}
\begin{aligned}
p(z_{1,N} \mid S ) &\propto \exp(\bu_{S}^\top \bu_{z_{1,N}}),&
p(z_{i,j}, z_{j,k} \mid z_{i,k}) &\propto \begin{cases}
  \exp(\bu_{z_{i,k}}^\top \bv_{z_{i,j}\ z_{j,k}}) & \substack{i+1 = j \lor \\j+1=k} \\ 
 \phi(\bu_{z_{i,k}}')^\top \phi(\bv_{z_{i,j}), z_{j,k}} &\text{o.w.} \\
\end{cases}\\
 p(x_i \mid z_i) &\propto \exp(\bu_{z_i}^\top \bv_{x_i}),\\
\end{aligned}
\end{equation}
where $\bu_{z}$/$\bu_z'$ is the embedding of $z$ when $z$ is used as head,$\bv_{x}$/$\bv_{z_1, z_2}$ is the embedding of $x$/$(z_1, z_2)$ when they are used as tail. See Appendix~\ref{sec:mlp-param} for the full
parameterization, drawn from \citet{kim2019cpcfg}. Note that we limit
the application of low-rank constraints to nonterminal to nonterminal
productions. These productions dominate the runtime as they
are applied at $O(T^3)$ hyperedges.
This allows inference to be run in time $O(T^3L^2N)$, where $T$ is the length of a sequence, $L$ the size of the label space, and $N$ the feature dimension.

\section{Experimental Setup}
\label{sec:experiments}
We evaluate the application of low-rank constraints with four experiments: sequential language modeling with HMMs, polyphonic music modeling with a large observation space, hierarchical language modelings with PCFGs, and video modeling with HSMMs.

\textbf{Data}
Our first set of experiments evaluate sequential models on \textsc{Penn Treebank} dataset (\textsc{Ptb}) \citep{ptb} for the task of word-level language modeling.
We use the preprocessing from \citet{mikolov-2011}. The second set of experiments is on polyphonic music modeling \citep{polyphonic}.
We evaluate on four music datasets: Nottingham (Nott), Piano, MuseData (Muse), and JSB chorales (JSB).
Each timestep consists of an 88-dimensional binary vector indicating whether a particular note is played. Since multiple notes may be played at the same time, the effective vocabulary size is extremely large. The third set of experiments use PCFGs for language modeling, we also use \textsc{Ptb}, but with the splits and preprocessing used in unsupervised constituency parsing \citep{shen2018prpn,shen2018ordered,kim2019cpcfg}. The last set of experiments use HSMMs for video modeling, where we use \textsc{CrossTask} \citep{zhukov2019cross} with 10\% of the training data for validation. We follow the preprocessing steps in \citet{fried2020learning} and apply PCA to project features to vectors of size 200. For the full details on datasets, please see Appendix~\ref{sec:data}.

\textbf{Models and Hyperparameters}
For language modeling with HMMs, we experiment with a range of state sizes, $|\mcL| = L \in \set{2^{10}, 2^{11}, 2^{12}, 2^{13}, 2^{14}}$,
and rank $N \in \set{L/2, L/4, L/8}$.
For polyphonic music modeling with HMMs, we experiment with states sizes $L \in \set{2^7, 2^8, 2^9, 2^{10}, 2^{11}}$.
For language modeling with PCFGs, we use a set of nonterminals of size $|\mcN| \in \set{30, 60, 100}$ and preterminals of twice the number of nonterminals $|\mcP|=2|\mcN|$. Our smallest setting ($|\mcN|=30$, $|\mcP|=60$) is the one used in \citet{kim2019cpcfg}. For video modeling with HSMMs, we use the same model setting as \citet{fried2020learning}, but we don't constrain states to the predefined states per task, and we experiment with state sizes $L \in \set{2^6, 2^7, 2^8, 2^9, 2^{10}}$ and rank $N \in \set{2^4, 2^5, 2^6, 2^7}$.

We utilize the feature map $\phi(x) = \exp(Wx)$ for the LHMM and LHSMM, and $\phi(x) = \exp(Wx - \|x\|_2^2/2)$ for the LPCFG. We initialize the parameters of feature maps using orthogonal feature projections \citep{choromanski2020performer}, and update it alongside the model parameters.
For the full hyperparameter and optimization details, see Appendix~\ref{sec:opt}.

\textbf{Baselines and Evaluation}
The language modeling experiments are evaluated using perplexity. Baselines are neurally parameterized HMM
with a standard softmax transition.  We also compare to VL-HMM, which makes a strong structural sparsity assumption on the emission distribution~\citep{chiu2020scaling}. We include for reference a state-of-the-art language model, the AWD-LSTM~\citep{merity2017awdlstm}. For polyphonic music modeling, we compare our LHMM against RNN-NADE~\citep{polyphonic} which models the full joint distribution of notes as well as temporal dependencies; as well as autoregressive neural models such as the R-Transformer~\citep{rtransformer} (as reported by \citet{betalstm}) and an LSTM ~(as reported by \citet{flow}); models with latent continuous dynamics such as the LV-RNN \citep{nasmc} and SRNN \citep{srnn}; and finally comparable models with latent discrete dynamics, the TSBN \citep{tsbn} and the baseline HMM.
We evaluate perplexities of our low-rank PCFG (LPCFG) against a softmax PCFG (PCFG) \citep{kim2019cpcfg}. For video modeling, we evaluate the negative log likelihoods on the test set and compare low-rank HSMMs to softmax HSMMs.

\section{Results}\label{sec:results}

\textbf{Hidden Markov Models for Language Modeling}
\begin{figure}[t]
\centering
\begin{subfigure}[t]{0.40\textwidth}
\centering
\includegraphics[height=4.5cm,trim={0 0 6.5cm 0}, clip]{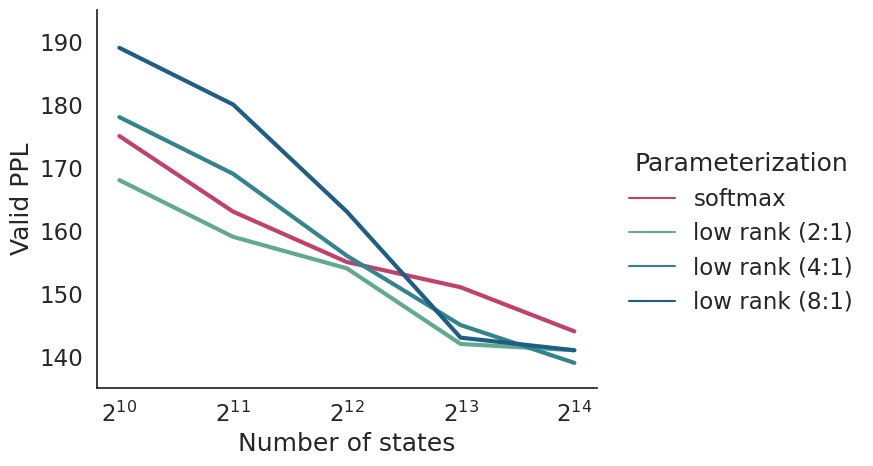}
\end{subfigure}
\hspace{1.25em}
\begin{subfigure}[t]{0.50\textwidth}
\centering
\includegraphics[height=4.5cm]{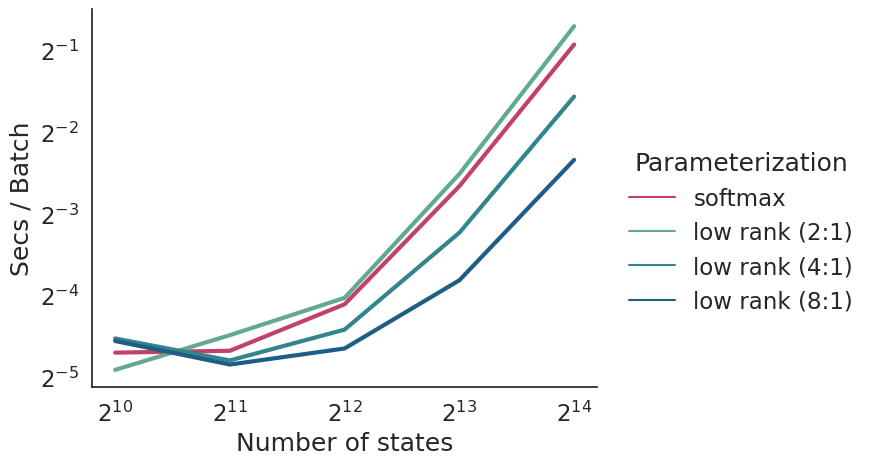}
\end{subfigure}
\caption{
\label{fig:hmm-ppl-features-dropout}
Validation perplexities on \textsc{Ptb} versus model scale, as well as speed in seconds per batch.
}
\end{figure}
Our main experimental result is that the low-rank models achieve similar accuracy,
as measured by perplexity, as our baselines.
Fig.~\ref{fig:hmm-ppl-features-dropout} shows that perplexity improves as we increase the scale of the HMM, and
that the performance of our LHMM also improves at the same rate.
At small sizes, the low-rank constraints slightly hinder accuracy; however
once the size is large enough, i.e. larger than $2^{12}$, LHMMs with 8:1 state-to-rank ratios perform comparably. \footnote{
See Appendix~\ref{sec:rank} for an analysis of the ranks of HMMs/LHMMs.}

Fig.~\ref{fig:hmm-ppl-features-dropout} also contains speed comparisons between HMMs and LHMMs.
A state-to-rank ratio of 8:1 matches the accuracy of softmax HMMs at larger state sizes and also gives an empirical speedup of more than 3x at $L = 2^{14}$.
As expected, we only see a speedup when the state-to-rank ratio exceeds 2:1,
as we replaced the $O(L^2)$ operation with two $O(LN)$ ones.
This implies that the low-rank constraint is most effective with scale,
where we observe large computational gains at no cost in accuracy.

\begin{table}[!t]
\centering
\begin{minipage}[t]{0.4\textwidth}
\centering
\begin{tabular}{lrr}
\toprule
Model & Val & Test\\

\midrule
AWD-LSTM & 60.0 & 57.3\\
VL-HMM & 128.6 & 119.5\\
HMM & 144.3 & 136.8\\
LHMM & 141.4 & 131.8\\
\bottomrule
\end{tabular}
\end{minipage}
\vspace{0em}
\hfill
\begin{minipage}[t]{0.55\textwidth}
\centering
\begin{tabular}{lrrr}
\toprule
Model & $L:N$ & Train & Val\\
\midrule
HMM & -  & 95.9 & 144.3\\
LHMM & 8 & 97.5 & 141.4\\
LHMM+band & 8 & 101.1 & 143.8\\
LHMM & 16 & 110.6 & 146.3 \\
LHMM+band & 16 & 96.9 & 138.8\\
LHMM & 32 & 108.4 & 153.7 \\
LHMM+band & 32 & 110.7 & 145.0\\
\bottomrule
\end{tabular}
\end{minipage}
\vspace{0.5em}
\caption{\label{tbl:hmm}
Model perplexities on \textsc{Ptb}.
All HMM variants have $L=2^{14}$ states.
(Left): Validation and test perplexities.
The LHMM has a state-to-rank ratio $8:1$.
(Right): Further experiments with extending the low-rank structure of LHMMs with a banded transition structure.
}
\end{table}


HMMs are outperformed by neural models,
and also by VL-HMMs \citep{chiu2020scaling} which offer similar modeling advantages to HMMs,
as shown in Tbl.~\ref{tbl:hmm}~(left).
This indicates that some aspects of performance are not strictly tied to scale. 
We posit this is due to the problem-specific block-sparse emission constraint in VL-HMMs. While 
very effective for language modeling,
the VL-HMM relies on a hard clustering of states for constraining emissions. 
This is difficult to apply to problems with richer emission models (as in music and video modeling).

\begin{figure}[t]
\centering
\begin{subfigure}{0.45\textwidth}
\centering
\includegraphics[height=4.5cm,trim={0 0 0em 0},clip]{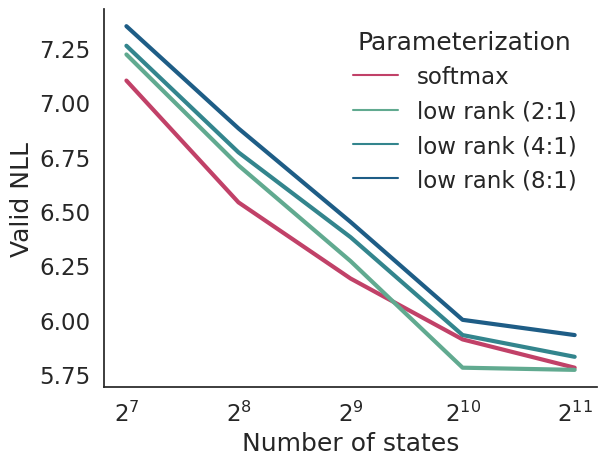}
\end{subfigure}
\hspace{0.75em}
\begin{subfigure}{0.50\textwidth}
\centering
\begin{tabular}{lrrrr}
\toprule
Model       & Nott & Piano & Muse & JSB \\
\midrule
RNN-NADE & 2.31  & \textbf{7.05}        & \textbf{5.6}        & 5.19          \\
R-Transformer & \textbf{2.24} & 7.44 & 7.00 & 8.26 \\
LSTM  & 3.43 & 7.77   & 7.23 & 8.17     \\
LV-RNN    & 2.72       & 7.61        & 6.89       &\textbf{ 3.99}\\
SRNN     & 2.94       & 8.20         & 6.28       & 4.74          \\
\midrule
TSBN     & 3.67       & 7.89        & 6.81       & 7.48          \\
HMM &  2.43 & 8.51 & 7.34 & 5.74 \\
LHMM & 2.60 & 8.89 & 7.60 & 5.80 \\
\bottomrule
\end{tabular}
\end{subfigure}
\caption{\label{fig:music}
Polyphonic music negative log-likelihoods (NLL), measured in nats.
(Left): HMM and LHMM validation performance for various state sizes and state:rank ratios. 
(Right): Test-set NLLs for polyphonic music.
The HMM models have $\mcL=2^{11}$ states and the LHMM has rank $N=2^{9}$, a 4:1 state:rank ratio.}
\end{figure}

\textbf{Hidden Markov Models for Music Modeling}
We next apply LHMMs on polyphonic music modeling. This has a max effective vocabulary size of $2^{88}$, as multiple notes may occur simultaneously. Unlike for language modeling, we use a factored Bernoulli emission model, modeling the presence of each note independently. Fig.~\ref{fig:music} (right) shows that HMMs are competitive with many of the models on these datasets, including LSTMs.
We find that LHMMs achieve performance slightly worse than but comparable to the unconstrained HMMs overall.
Fig.~\ref{fig:music} (left) shows that the distinction drops with more states. Both HMMs achieve low negative likelihoods (NLL) on the datasets with shorter sequences, Nottingham and JSB, but relatively poorer NLLs on the datasets with longer sequences (Muse and Piano).

\textbf{Context-Free Grammars}
For syntactic language modeling on \textsc{Ptb}, our low-rank PCFG (LPCFG) achieves similar performance to PCFGs, as shown in Table~\ref{tbl:cky} (left), with an improvement in computational complexity. The complexity of inference in PCFGs models is cubic in the number of nonterminals, so even models with $|\mcN|=30$ nonterminals are relatively
costly. Our approach achieves comparable results with $N=8$ features. As we scale up the number of nonterminals to $|\mcN| = 100$, LPCFG stays competitive with a lower computational complexity (since $N < |\mcN|$). These experiments also demonstrate the importance of scale in syntactic language models with more than 50 point gain in perplexity over a strong starting model.

\textbf{CFG Speed}
Once the model is large enough, i.e. $|\mathcal{N}|\ge 60$ nonterminals and
$|\mathcal{P}|\ge 120$ preterminals, the LPCFG is faster than PCFG,
as shown in Tbl.~\ref{tbl:cky} (left).
Note that the LPCFG is faster than the CFG even when the number of features
$N>\frac{|\mathcal{N}|}{2}$, in contrast to the HMM case where a speedup can only
be obtained when $N < L/2$.
This is due to the scoring matrix being rectangular:
Recall the low-rank matrix product
$\Psi\beta = U(V^\top\beta)$, where,
when specialized to PCFGs, the left-hand side takes time $O(L^{3})$
and the right-hand side takes $O(L^{2}N + LN)$.
For PCFGs, the term $V^\top\beta$ dominates the runtime.
This contrasts with HMMs, where both $V^\top\beta$ and the subsequent multiplication
by $U$ take the same amount of time, $O(LN)$.

\textbf{Hidden Semi-Markov Models for Video Modeling} Table~\ref{tbl:cky} (right)
shows the results of video modeling using HSMMs.
In addition to using a different hypergraph for inference,
these experiments use a continuous Gaussian emission model.
By removing the state constraints from tasks,
our HSMM baselines get better video-level NLLs than that from \citet{fried2020learning}
at the cost of more memory consumption.
Due to GPU memory constraints, we can only train HSMMs up to $2^8$ states.
However, the low-rank parameterization allows models to scale to $2^{10}$ states,
yielding an improvement in NLL.
Absolute results could likely be improved with more states and by an
improved emission parameterization for all models.

\textbf{Improving Rank Assumptions}
One potential limitation of all low-rank models is that they cannot learn high-rank structures with low $N$.
We began to see this issue at a ratio of 16:1 states to features for large HMMs.
To explore the effects of this limitation, we perform an additional experiment that combines low-rank features with a sparse component. Specifically we add an efficient high-rank sparse banded transition matrix. 
The full details are in Appendix~\ref{sec:banded}.
Tbl.~\ref{tbl:hmm} (right) shows that combination with the band structure allows for larger ratios than just the low-rank structure alone, while only adding another operation that costs $O(LN)$.

\begin{table}[!t]
\centering
\begin{tabular} {lllrrrr}
\toprule
$|\mathcal{N}|$ & $|\mathcal{P}|$ & Model & $N$ &  PPL & Secs\\ 
\midrule
30  & 60    & PCFG & - & 252.60 &  0.23\\  
    &       & LPCFG & 8 &  247.02    & 0.27\\ 
    &       & LPCFG & 16 & 250.59    & 0.27\\ 
\midrule
60  & 120   & PCFG & - & 234.01 &  0.33\\ 
    &       & LPCFG & 16& 217.24 & 0.28\\ 
    &       & LPCFG & 32& 213.81 & 0.30\\ 
\midrule
100 & 200   & PCFG & - &  191.08   & 1.02\\ 
    &       & LPCFG & 32& 203.47 & 0.64\\ 
    &       & LPCFG & 64& 194.25 & 0.81\\ 
\bottomrule
\end{tabular}
\vspace{0em}
\hfill
\begin{tabular} {lllrrrr}
\toprule
Model & $L$ & $N$ & NLL & Secs \\ 
\midrule
HSMM\footnote{We report the NLL of the baseline model taken from \citet{fried2020learning}, where every label corresponds to one of 151 possible actions.} & 151 & - & $1.432e5$ & -\\ 
\midrule
HSMM & $2^6$ & - & $1.428e5$ & 0.78\\ 
HSMM & $2^7$ & - & $1.427e5$  & 2.22\\ 
HSMM & $2^8$ & - & $1.426e5$ & 7.69\\ 
\midrule
LHSMM & $2^7$ & $2^7$ & $1.427e5$ & 4.17\\ 
LHSMM & $2^8$ & $2^6$ & $1.426e5$ & 5.00\\ 
LHSMM & $2^9$ & $2^5$ & $1.424e5$ & 5.56\\ 
LHSMM & $2^{10}$ & $2^4$ & $1.423e5$ & 10.0\\ 
\bottomrule
\end{tabular}
\vspace{.5em}
\caption{\label{tbl:cky}
(Left): Test perplexities and speeds for PCFG models on \textsc{Ptb}.
The complexity of PCFG is $O(T^3|\mcN|^3)$, whereas the complexity of LPCFG is $O(T^3 |\mcN|^2N)$.
Speeds are given in seconds per batch. 
(Right): Negative log likelihoods (NLL) per video and speeds for HSMM models on \textsc{CrossTask}.
We cannot train HSMMs beyond $2^8$ states due to GPU memory constraints,
but we can train LHSMMs with up to $2^{10}$ states.
Speeds are given in seconds per batch. 
}
\end{table}

\section{Related Work}

Similar to our work, other approaches target matrix or tensor operations in inference, and impose structural model constraints to improve computational complexity.
Many of the works on HMMs in particular take advantage of the transition structure.
The Dense-mostly-constant (DMC) HMM assigns a subset of learnable parameters per row of the transition matrix and sets the rest to a constant, leading to a sub-quadratic runtime \citep{dmc}.
Other structures have also been explored, such as aligning the states of an HMM to underlying phenomena that allows inference to be sped up \citep{ffthmm,constrainedhmm}.
Additionally, other methods take advantage of emission structure in HMMs in order to scale, such as the Cloned HMM \citep{dedieu2019learning} and VL-HMM \citep{chiu2020scaling}.
Compared to these approaches, our method is more flexible and generic, since it can be applied in a non-application-specific manner, and even extended with high-rank components (such as banded structure).

Low-rank structure has been explored in both HMMs~\citep{rrhmm}, a generalization of PCFGs called weighted tree automata~\citep{rrpcfg}, and conditional random fields~\citep{thai2018embedded}. The reduced-rank HMM~\citep{rrhmm} has at most 50 states, and relies on spectral methods for training.  The low-rank weighted tree automata~\citep{rrpcfg} also trains latent tree models via spectral methods. We extend the low-rank assumption to neural parameterizations, which have been shown to be effective for generalization \citep{kim2019cpcfg,chiu2020scaling}, and directly optimize the evidence via gradient descent.
Finally, \citet{thai2018embedded} do not take advantage of the low-rank parameterization of their CRF potentials for faster inference via low-rank matrix products, a missed opportunity.
Instead, the low-rank parameterization is used only as a regularizer, with the full potentials instantiated during inference.

Concurrent work in unsupervised parsing uses a tensor decomposition to scale PCFGs to large state spaces \citep{yang2021pcfgs}. Our low-rank decomposition of the flattened head-tail scoring matrix is more general, resulting in worse scaling for the PCFG setting but with wider applicability, as shown by experiments with HMMs and HSMMs.

\section{Conclusion}

This work improves the scaling of structured models by establishing the effectiveness of low-rank constraints for hypergraph models. We show that viewing a key step of inference in structured models as a matrix-vector product, in combination with a low-rank constraint on relevant parameters, allows for an immediate speedup. 
Low-rank inference allows us to obtain a reduction in the asymptotic complexity of marginalization at the cost of 
a constrained model. 
Our approach applies to a wide class of models, including HMMs, HSMMs, and PCFGs. Through our experiments on language, video, and polyphonic music modeling, we demonstrate an effective approach for overcoming the practical difficulty of applying low-rank constraints in high dimensional, structured spaces by targeting and constraining model components that bottleneck computation. Future work includes exploration of other structural constraints for speeding up matrix-vector products~\citep{kaleidoscope} performed in inference, as well as application to models where exact inference is intractable.

\begin{ack}
We thank Nikita Kitaev for the discussion that sparked this project.
We thank Sam Wiseman, Jack Morris, and the anonymous reviewers for valuable feedback.
Yuntian Deng is sponsored by NSF 1704834, Alexander Rush by NSF CAREER 1845664, and Justin Chiu by an Amazon research award.

\end{ack}

\bibliography{bib}

\begin{thebibliography}{49}
\providecommand{\natexlab}[1]{#1}
\providecommand{\url}[1]{\texttt{#1}}
\expandafter\ifx\csname urlstyle\endcsname\relax
  \providecommand{\doi}[1]{doi: #1}\else
  \providecommand{\doi}{doi: \begingroup \urlstyle{rm}\Url}\fi

\bibitem[Bayer and Osendorfer(2015)]{storn}
Justin Bayer and Christian Osendorfer.
\newblock Learning stochastic recurrent networks, 2015.

\bibitem[Bengio et~al.(2003)Bengio, Ducharme, Vincent, and Janvin]{nplm}
Yoshua Bengio, R\'{e}jean Ducharme, Pascal Vincent, and Christian Janvin.
\newblock A neural probabilistic language model.
\newblock \emph{J. Mach. Learn. Res.}, 3:\penalty0 1137–1155, March 2003.
\newblock ISSN 1532-4435.

\bibitem[Boulanger-Lewandowski et~al.(2012)Boulanger-Lewandowski, Bengio, and
  Vincent]{polyphonic}
Nicolas Boulanger-Lewandowski, Yoshua Bengio, and Pascal Vincent.
\newblock Modeling temporal dependencies in high-dimensional sequences:
  Application to polyphonic music generation and transcription.
\newblock In \emph{Proceedings of the 29th International Coference on
  International Conference on Machine Learning}, ICML'12, page 1881–1888,
  Madison, WI, USA, 2012. Omnipress.
\newblock ISBN 9781450312851.

\bibitem[Brown et~al.(2020)Brown, Mann, Ryder, Subbiah, Kaplan, Dhariwal,
  Neelakantan, Shyam, Sastry, Askell, Agarwal, Herbert-Voss, Krueger, Henighan,
  Child, Ramesh, Ziegler, Wu, Winter, Hesse, Chen, Sigler, Litwin, Gray, Chess,
  Clark, Berner, McCandlish, Radford, Sutskever, and Amodei]{gpt3}
Tom~B. Brown, Benjamin Mann, Nick Ryder, Melanie Subbiah, Jared Kaplan,
  Prafulla Dhariwal, Arvind Neelakantan, Pranav Shyam, Girish Sastry, Amanda
  Askell, Sandhini Agarwal, Ariel Herbert-Voss, Gretchen Krueger, Tom Henighan,
  Rewon Child, Aditya Ramesh, Daniel~M. Ziegler, Jeffrey Wu, Clemens Winter,
  Christopher Hesse, Mark Chen, Eric Sigler, Mateusz Litwin, Scott Gray,
  Benjamin Chess, Jack Clark, Christopher Berner, Sam McCandlish, Alec Radford,
  Ilya Sutskever, and Dario Amodei.
\newblock Language models are few-shot learners, 2020.

\bibitem[Chiang and Riley(2020)]{chiang2020factor}
David Chiang and Darcey Riley.
\newblock Factor graph grammars.
\newblock \emph{arXiv preprint arXiv:2010.12048}, 2020.

\bibitem[Chiu and Rush(2020)]{chiu2020scaling}
Justin Chiu and Alexander Rush.
\newblock Scaling hidden {M}arkov language models.
\newblock In \emph{Proceedings of the 2020 Conference on Empirical Methods in
  Natural Language Processing (EMNLP)}, pages 1341--1349, Online, November
  2020. Association for Computational Linguistics.
\newblock \doi{10.18653/v1/2020.emnlp-main.103}.
\newblock URL \url{https://www.aclweb.org/anthology/2020.emnlp-main.103}.

\bibitem[Choromanski et~al.(2020)Choromanski, Likhosherstov, Dohan, Song, Gane,
  Sarlos, Hawkins, Davis, Mohiuddin, Kaiser, Belanger, Colwell, and
  Weller]{choromanski2020performer}
Krzysztof Choromanski, Valerii Likhosherstov, David Dohan, Xingyou Song,
  Andreea Gane, Tamas Sarlos, Peter Hawkins, Jared Davis, Afroz Mohiuddin,
  Lukasz Kaiser, David Belanger, Lucy Colwell, and Adrian Weller.
\newblock Rethinking attention with performers, 2020.

\bibitem[Dao et~al.(2020)Dao, Sohoni, Gu, Eichhorn, Blonder, Leszczynski,
  Rudra, and Ré]{kaleidoscope}
Tri Dao, Nimit Sohoni, Albert Gu, Matthew Eichhorn, Amit Blonder, Megan
  Leszczynski, Atri Rudra, and Christopher Ré.
\newblock Kaleidoscope: An efficient, learnable representation for all
  structured linear maps.
\newblock In \emph{International Conference on Learning Representations}, 2020.
\newblock URL \url{https://openreview.net/forum?id=BkgrBgSYDS}.

\bibitem[Dedieu et~al.(2019)Dedieu, Gothoskar, Swingle, Lehrach,
  Lázaro-Gredilla, and George]{dedieu2019learning}
Antoine Dedieu, Nishad Gothoskar, Scott Swingle, Wolfgang Lehrach, Miguel
  Lázaro-Gredilla, and Dileep George.
\newblock Learning higher-order sequential structure with cloned hmms, 2019.

\bibitem[Felzenszwalb et~al.(2004)Felzenszwalb, Huttenlocher, and
  Kleinberg]{ffthmm}
Pedro Felzenszwalb, Daniel Huttenlocher, and Jon Kleinberg.
\newblock Fast algorithms for large-state-space hmms with applications to web
  usage analysis.
\newblock In S.~Thrun, L.~Saul, and B.~Sch\"{o}lkopf, editors, \emph{Advances
  in Neural Information Processing Systems}, volume~16. MIT Press, 2004.
\newblock URL
  \url{https://proceedings.neurips.cc/paper/2003/file/9407c826d8e3c07ad37cb2d13d1cb641-Paper.pdf}.

\bibitem[Fraccaro et~al.(2016)Fraccaro, Sønderby, Paquet, and Winther]{srnn}
Marco Fraccaro, Søren~Kaae Sønderby, Ulrich Paquet, and Ole Winther.
\newblock Sequential neural models with stochastic layers, 2016.

\bibitem[Fried et~al.(2020)Fried, Alayrac, Blunsom, Dyer, Clark, and
  Nematzadeh]{fried2020learning}
Daniel Fried, Jean-Baptiste Alayrac, Phil Blunsom, Chris Dyer, Stephen Clark,
  and Aida Nematzadeh.
\newblock Learning to segment actions from observation and narration.
\newblock \emph{arXiv preprint arXiv:2005.03684}, 2020.

\bibitem[Gan et~al.(2015)Gan, Li, Henao, Carlson, and Carin]{tsbn}
Zhe Gan, Chunyuan Li, Ricardo Henao, David Carlson, and Lawrence Carin.
\newblock Deep temporal sigmoid belief networks for sequence modeling, 2015.

\bibitem[Glorot and Bengio(2010)]{glorot2010understanding}
Xavier Glorot and Yoshua Bengio.
\newblock Understanding the difficulty of training deep feedforward neural
  networks.
\newblock In \emph{Proceedings of the thirteenth international conference on
  artificial intelligence and statistics}, pages 249--256. JMLR Workshop and
  Conference Proceedings, 2010.

\bibitem[Gu et~al.(2015)Gu, Ghahramani, and Turner]{nasmc}
Shixiang Gu, Zoubin Ghahramani, and Richard~E. Turner.
\newblock Neural adaptive sequential monte carlo, 2015.

\bibitem[Huang et~al.(2018)Huang, Vaswani, Uszkoreit, Shazeer, Hawthorne, Dai,
  Hoffman, and Eck]{music-trans}
Cheng{-}Zhi~Anna Huang, Ashish Vaswani, Jakob Uszkoreit, Noam Shazeer, Curtis
  Hawthorne, Andrew~M. Dai, Matthew~D. Hoffman, and Douglas Eck.
\newblock An improved relative self-attention mechanism for transformer with
  application to music generation.
\newblock \emph{CoRR}, abs/1809.04281, 2018.
\newblock URL \url{http://arxiv.org/abs/1809.04281}.

\bibitem[Huang and Chiang(2005)]{huang2005better}
Liang Huang and David Chiang.
\newblock Better k-best parsing.
\newblock In \emph{Proceedings of the Ninth International Workshop on Parsing
  Technology}, pages 53--64, 2005.

\bibitem[Javidian et~al.(2020)Javidian, Wang, Lu, and
  Valtorta]{javidian2020hypergraph}
Mohammad~Ali Javidian, Zhiyu Wang, Linyuan Lu, and Marco Valtorta.
\newblock On a hypergraph probabilistic graphical model.
\newblock \emph{Annals of Mathematics and Artificial Intelligence}, 88\penalty0
  (9):\penalty0 1003--1033, 2020.

\bibitem[Katharopoulos et~al.(2020)Katharopoulos, Vyas, Pappas, and
  Fleuret]{katharopoulos2020lineartransformer}
A.~Katharopoulos, A.~Vyas, N.~Pappas, and F.~Fleuret.
\newblock Transformers are rnns: Fast autoregressive transformers with linear
  attention.
\newblock In \emph{Proceedings of the International Conference on Machine
  Learning (ICML)}, 2020.

\bibitem[Kim et~al.(2019)Kim, Dyer, and Rush]{kim2019cpcfg}
Yoon Kim, Chris Dyer, and Alexander Rush.
\newblock Compound probabilistic context-free grammars for grammar induction.
\newblock In \emph{Proceedings of the 57th Annual Meeting of the Association
  for Computational Linguistics}, pages 2369--2385, Florence, Italy, July 2019.
  Association for Computational Linguistics.
\newblock \doi{10.18653/v1/P19-1228}.
\newblock URL \url{https://www.aclweb.org/anthology/P19-1228}.

\bibitem[Kingma and Ba(2017)]{kingma2017adam}
Diederik~P. Kingma and Jimmy Ba.
\newblock Adam: A method for stochastic optimization, 2017.

\bibitem[Klein and Manning(2004)]{klein2004parsing}
Dan Klein and Christopher~D Manning.
\newblock Parsing and hypergraphs.
\newblock In \emph{New developments in parsing technology}, pages 351--372.
  Springer, 2004.

\bibitem[Koller and Friedman(2009)]{pgm}
Daphne Koller and Nir Friedman.
\newblock \emph{Probabilistic Graphical Models: Principles and Techniques -
  Adaptive Computation and Machine Learning}.
\newblock The MIT Press, 2009.
\newblock ISBN 0262013193.

\bibitem[Krishnan et~al.(2016)Krishnan, Shalit, and Sontag]{dmm}
Rahul~G. Krishnan, Uri Shalit, and David Sontag.
\newblock Structured inference networks for nonlinear state space models, 2016.

\bibitem[Levine(2018)]{levine2018reinforcement}
Sergey Levine.
\newblock Reinforcement learning and control as probabilistic inference:
  Tutorial and review, 2018.

\bibitem[Liu et~al.(2018)Liu, Zhuo, Du, Zhang, Zhu, and Guizani]{adv-phmm}
Xiaolei Liu, Zhongliu Zhuo, Xiaojiang Du, Xiaosong Zhang, Qingxin Zhu, and
  Mohsen Guizani.
\newblock Adversarial attacks against profile hmm website fingerprinting
  detection model.
\newblock \emph{Cognitive Systems Research}, 54, 12 2018.
\newblock \doi{10.1016/j.cogsys.2018.12.005}.

\bibitem[Loshchilov and Hutter(2017)]{adamw}
Ilya Loshchilov and Frank Hutter.
\newblock Fixing weight decay regularization in adam.
\newblock \emph{CoRR}, abs/1711.05101, 2017.
\newblock URL \url{http://arxiv.org/abs/1711.05101}.

\bibitem[Marcus et~al.(1993)Marcus, Marcinkiewicz, and Santorini]{ptb}
Mitchell~P. Marcus, Mary~Ann Marcinkiewicz, and Beatrice Santorini.
\newblock Building a large annotated corpus of english: The penn treebank.
\newblock \emph{Comput. Linguist.}, 19\penalty0 (2):\penalty0 313–330, June
  1993.
\newblock ISSN 0891-2017.

\bibitem[Merity et~al.(2017)Merity, Keskar, and Socher]{merity2017awdlstm}
Stephen Merity, Nitish~Shirish Keskar, and Richard Socher.
\newblock Regularizing and optimizing {LSTM} language models.
\newblock \emph{CoRR}, abs/1708.02182, 2017.
\newblock URL \url{http://arxiv.org/abs/1708.02182}.

\bibitem[Mikolov et~al.(2011)Mikolov, Deoras, Kombrink, Burget, and
  Cernocký]{mikolov-2011}
Tomas Mikolov, Anoop Deoras, Stefan Kombrink, Lukás Burget, and Jan Cernocký.
\newblock Empirical evaluation and combination of advanced language modeling
  techniques.
\newblock pages 605--608, 2011.
\newblock URL
  \url{http://dblp.uni-trier.de/db/conf/interspeech/interspeech2011.html#MikolovDKBC11}.

\bibitem[Peng et~al.(2021)Peng, Pappas, Yogatama, Schwartz, Smith, and
  Kong]{peng2021rfa}
Hao Peng, Nikolaos Pappas, Dani Yogatama, Roy Schwartz, Noah Smith, and
  Lingpeng Kong.
\newblock Random feature attention.
\newblock In \emph{International Conference on Learning Representations}, 2021.
\newblock URL \url{https://openreview.net/forum?id=QtTKTdVrFBB}.

\bibitem[Rabusseau et~al.(2015)Rabusseau, Balle, and Cohen]{rrpcfg}
Guillaume Rabusseau, Borja Balle, and Shay~B. Cohen.
\newblock Weighted tree automata approximation by singular value truncation.
\newblock \emph{CoRR}, abs/1511.01442, 2015.
\newblock URL \url{http://arxiv.org/abs/1511.01442}.

\bibitem[Roweis(2000)]{constrainedhmm}
Sam Roweis.
\newblock Constrained hidden markov models.
\newblock In S.~Solla, T.~Leen, and K.~M\"{u}ller, editors, \emph{Advances in
  Neural Information Processing Systems}, volume~12. MIT Press, 2000.
\newblock URL
  \url{https://proceedings.neurips.cc/paper/1999/file/84c6494d30851c63a55cdb8cb047fadd-Paper.pdf}.

\bibitem[Shen et~al.(2018)Shen, Lin, wei Huang, and Courville]{shen2018prpn}
Yikang Shen, Zhouhan Lin, Chin wei Huang, and Aaron Courville.
\newblock Neural language modeling by jointly learning syntax and lexicon.
\newblock In \emph{International Conference on Learning Representations}, 2018.
\newblock URL \url{https://openreview.net/forum?id=rkgOLb-0W}.

\bibitem[Shen et~al.(2019)Shen, Tan, Sordoni, and Courville]{shen2018ordered}
Yikang Shen, Shawn Tan, Alessandro Sordoni, and Aaron Courville.
\newblock Ordered neurons: Integrating tree structures into recurrent neural
  networks.
\newblock In \emph{International Conference on Learning Representations}, 2019.
\newblock URL \url{https://openreview.net/forum?id=B1l6qiR5F7}.

\bibitem[Siddiqi and Moore(2005)]{dmc}
Sajid~M. Siddiqi and Andrew~W. Moore.
\newblock Fast inference and learning in large-state-space hmms.
\newblock In \emph{Proceedings of the 22nd International Conference on Machine
  Learning}, ICML '05, page 800–807, New York, NY, USA, 2005. Association for
  Computing Machinery.
\newblock ISBN 1595931805.
\newblock \doi{10.1145/1102351.1102452}.
\newblock URL
  \url{https://doi-org.proxy.library.cornell.edu/10.1145/1102351.1102452}.

\bibitem[Siddiqi et~al.(2009)Siddiqi, Boots, and Gordon]{rrhmm}
Sajid~M. Siddiqi, Byron Boots, and Geoffrey~J. Gordon.
\newblock Reduced-rank hidden markov models.
\newblock \emph{CoRR}, abs/0910.0902, 2009.
\newblock URL \url{http://arxiv.org/abs/0910.0902}.

\bibitem[Song et~al.(2019)Song, Jang, Shin, and Moon]{betalstm}
Kyungwoo Song, JoonHo Jang, Seungjae Shin, and Il{-}Chul Moon.
\newblock Bivariate beta {LSTM}.
\newblock \emph{CoRR}, abs/1905.10521, 2019.
\newblock URL \url{http://arxiv.org/abs/1905.10521}.

\bibitem[Stoller et~al.(2019)Stoller, Tian, Ewert, and Dixon]{sequnet}
Daniel Stoller, Mi~Tian, Sebastian Ewert, and Simon Dixon.
\newblock Seq-u-net: {A} one-dimensional causal u-net for efficient sequence
  modelling.
\newblock \emph{CoRR}, abs/1911.06393, 2019.
\newblock URL \url{http://arxiv.org/abs/1911.06393}.

\bibitem[Thai et~al.(2018)Thai, Ramesh, Murty, Vilnis, and
  McCallum]{thai2018embedded}
Dung Thai, Sree~Harsha Ramesh, Shikhar Murty, Luke Vilnis, and Andrew McCallum.
\newblock Embedded-state latent conditional random fields for sequence
  labeling.
\newblock In \emph{Proceedings of the 22nd Conference on Computational Natural
  Language Learning}, pages 1--10, Brussels, Belgium, October 2018. Association
  for Computational Linguistics.
\newblock \doi{10.18653/v1/K18-1001}.
\newblock URL \url{https://aclanthology.org/K18-1001}.

\bibitem[Wang et~al.(2019)Wang, Ma, Liu, and Tang]{rtransformer}
Zhiwei Wang, Yao Ma, Zitao Liu, and Jiliang Tang.
\newblock R-transformer: Recurrent neural network enhanced transformer, 2019.

\bibitem[Weissenborn et~al.(2020)Weissenborn, Täckström, and
  Uszkoreit]{video}
Dirk Weissenborn, Oscar Täckström, and Jakob Uszkoreit.
\newblock Scaling autoregressive video models.
\newblock In \emph{International Conference on Learning Representations}, 2020.
\newblock URL \url{https://openreview.net/forum?id=rJgsskrFwH}.

\bibitem[Yang et~al.(2021)Yang, Zhao, and Tu]{yang2021pcfgs}
Songlin Yang, Yanpeng Zhao, and Kewei Tu.
\newblock Pcfgs can do better: Inducing probabilistic context-free grammars
  with many symbols, 2021.

\bibitem[Yu(2010)]{yu2010hidden}
Shun-Zheng Yu.
\newblock Hidden semi-markov models.
\newblock \emph{Artificial intelligence}, 174\penalty0 (2):\penalty0 215--243,
  2010.

\bibitem[Zhang et~al.(2021)Zhang, Cheng, Zhang, and Li]{Zhang2021LabelFA}
Hongpo Zhang, Ning Cheng, Yang Zhang, and Zhanbo Li.
\newblock Label flipping attacks against naive bayes on spam filtering systems.
\newblock \emph{Applied Intelligence}, pages 1--12, 2021.

\bibitem[Zhang et~al.(2018)Zhang, Wang, Ji, Shen, and Wang]{adv-interp}
Xinyang Zhang, Ningfei Wang, Shouling Ji, Hua Shen, and Ting Wang.
\newblock Interpretable deep learning under fire.
\newblock \emph{CoRR}, abs/1812.00891, 2018.
\newblock URL \url{http://arxiv.org/abs/1812.00891}.

\bibitem[Zhou et~al.(2006)Zhou, Huang, and Sch{\"o}lkopf]{zhou2006learning}
Dengyong Zhou, Jiayuan Huang, and Bernhard Sch{\"o}lkopf.
\newblock Learning with hypergraphs: Clustering, classification, and embedding.
\newblock \emph{Advances in neural information processing systems},
  19:\penalty0 1601--1608, 2006.

\bibitem[Zhukov et~al.(2019)Zhukov, Alayrac, Cinbis, Fouhey, Laptev, and
  Sivic]{zhukov2019cross}
Dimitri Zhukov, Jean-Baptiste Alayrac, Ramazan~Gokberk Cinbis, David Fouhey,
  Ivan Laptev, and Josef Sivic.
\newblock Cross-task weakly supervised learning from instructional videos.
\newblock In \emph{Proceedings of the IEEE/CVF Conference on Computer Vision
  and Pattern Recognition}, pages 3537--3545, 2019.

\bibitem[Ziegler and Rush(2019)]{flow}
Zachary~M. Ziegler and Alexander~M. Rush.
\newblock Latent normalizing flows for discrete sequences, 2019.

\end{thebibliography}
\bibliographystyle{plainnat}

\newpage

\appendix

\section{\label{sec:expressivity}Expressivity of Low-Rank Models}
We focus on the simplest case of HMMs for an analysis of expressivity.
In the case of Gaussian emissions, a model with more states but low rank is more expressive than a model with fewer states because for a single timestep, a larger mixture of Gaussians is more expressive.
In the case of discrete emissions, however, the emission distribution for a single timestep (i.e. $\sum_{z} p(x, z)$) is not more expressive. Instead, we show that there exists joint marginal distributions of discrete $x$ over multiple timesteps that are captured by large state but low-rank HMMs, but not expressible by models with fewer states.

We construct a counter-example with a sequence of length $T=2$ and emission space of $x_t \in \set{0,1,2}$.
We show that a 3-state HMM with rank 2, HMM-3-2, with manually chosen transitions and emissions, cannot be modeled by any 2-state HMM.
The transition probabilities for the HMM-3-2 are given by (rows $z_t$, columns $z_{t+1}$)
$$
p(z_{t+1} |z_{t})
= \begin{bmatrix}
    \frac{1}{3} & \frac{1}{3} & \frac{1}{3} \\
    0 & 1 & 0 \\
    \frac{1}{2} & 0 & \frac{1}{2}
\end{bmatrix}
=  \begin{bmatrix}
    \frac{1}{3} & \frac{2}{3} \\
    1 & 0 \\
    0 & 1
\end{bmatrix}
\begin{bmatrix}
    0 & 1 & 0 \\
    \frac{1}{2} & 0 & \frac{1}{2}
\end{bmatrix} = UV^T,
$$
emission probabilities by (rows $z_t$, columns $x_t$):
$$p(x_t|z_t) = \begin{bmatrix}1 &0 & 0 \\ 0 & 1 & 0 \\ 0 & 0 &1\end{bmatrix},$$
and starting distribution
$$P(z_1\mid z_0) = \begin{bmatrix}\frac{1}{3} & \frac{1}{3} & \frac{1}{3} \end{bmatrix}.$$
This yields the following marginal distribution (row $x_1$, column $x_2$):
$$
p(x_1, x_2) = \begin{bmatrix}
    \frac{1}{9} & \frac{1}{9} & \frac{1}{9} \\
    0 & \frac{1}{3} & 0 \\
    \frac{1}{6} & 0 & \frac{1}{6}
\end{bmatrix}.
$$

Next, we show that there does not exist a 2-state HMM that can have this marginal distribution.
Assuming the contrary, that there exists a 2-state HMM that has this marginal distribution, we will first show that there is only one possible emission matrix. We will then use that to further show that the posterior, then transitions also must be sparse, resulting in a marginal emission distribution that contradicts the original assumption.

We start by setting up a system of equations. The marginal distribution over observations is obtained by summing over $z_1,z_2$:
$$p(x_1,x_2) = \sum_{z_2}\left(\sum_{z_1}p(x_1,z_1,z_2)\right)p(x_2\mid z_2).$$
Let the inner term be $f(x_1,z_1) = \sum_{z_1}p(x_1,z_1,z_2)=p(x_1,z_2)$.
In a small abuse of notation, let $p(x_2\mid z_2=0)$ be a row vector with entries $[p(x_2\mid z_2=0)]_{x} = p(x_2=x\mid z_2=0)$, and similarly for $p(x_2 \mid z_2=1)$.
We then have, first summing over $z_1$,
$$
P(x_1, x_2)
=  \begin{bmatrix}
    \frac{1}{9} & \frac{1}{9} & \frac{1}{9} \\
    0 & \frac{1}{3} & 0 \\
    \frac{1}{6} & 0 & \frac{1}{6}
\end{bmatrix}
= \begin{bmatrix} f(0, 0) & f(0, 1) \\ f(1, 0) & f(1, 1) \\ f(2, 0) & f(2, 1) \end{bmatrix}
\begin{bmatrix}
p(x_2|z_2=0) \\ 
p(x_2|z_2=1)
\end{bmatrix}.
$$

We can determine the first row of the emission matrix, $p(x_2\mid z_2=0)$ from the second row of this system of equations,
rewritten here:
$$
p(x_2 \mid x_1=1) = f(1,0) p(x_2|z_2=0) + f(1,1)p(x_2|z_2=1)= \begin{bmatrix}0 & \frac{1}{3} & 0 \end{bmatrix}.$$
We can deduce that $f(1,0),f(1,1)>0$, otherwise $p(x_1,x_2=1) = 0 \ne \frac{1}{3}$. Without loss of generality, assume $f(1,0)> 0$, then $p(x_2=0|z_2=0)=p(x_2=2|z_2=0)=0$, since $p(x_1=1, x_2=0) = p(x_1=1,x_2=2) = 0$. Therefore,
$$p(x_2|z_2=0)=\begin{bmatrix}0 & 1& 0 \end{bmatrix}.$$

We can similarly determine the second row of the emission matrix, $p(x_2\mid z_2=1)$, from the last row of the system of equations:
$$
p(x_2 \mid x_1=2) = f(2,0) p(x_2|z_2=0) + f(2,1)p(x_2|z_2=1)= \begin{bmatrix} \frac{1}{6} &0 &  \frac{1}{6} \end{bmatrix}.$$
As we determined that $p(x_2|z_2=0)=\begin{bmatrix}0 & 1& 0 \end{bmatrix}$, $f(2,0)$ must be 0, otherwise $p(x_2=1|x_1=2)>0$. Therefore $f(2,1)p(x_2|z_2=1)= \begin{bmatrix} \frac{1}{6} &0 &  \frac{1}{6} \end{bmatrix}$, yielding $$p(x_2|z_2=1)= \begin{bmatrix} \frac{1}{2} &0 &  \frac{1}{2} \end{bmatrix}.$$

Putting it together, the full emission matrix is given by
$$p(x_t|z_t)=\begin{bmatrix}0 & 1& 0\\  \frac{1}{2} &0 &  \frac{1}{2}  \end{bmatrix}.$$
This allows us to find the posterior distribution $p(z_1 \mid x_1)$ via Bayes' rule:
$$
    p(z_1=1 \mid x_1=1) = \frac{p(x_1=1\mid z_1=1)p(z_1)}{p(x_1=1)}=
    \frac{0\cdot p(z_1)}{p(x_1=1)} = 0,
$$
implying $p(z_1=0\mid x_1=1) = 1$.
By similar reasoning, we have $p(z_1=1\mid x_1=0) = 1$ and $p(z_1=1\mid x_1=2) = 1$.

Given the sparse emissions and posteriors, we will show that the transitions must be similarly sparse, resulting in an overly sparse marginal distribution over emissions (contradiction).
We can lower bound 
$$0 = p(x_2=1\mid x_1=2)\ge p(x_2=1\mid z_2=0)p(z_2=0\mid z_1=1)p(z_1=1\mid x_1=2)$$
by the definition of total probability and nonnegativity of probability.
Then, substituting $p(x_2=1 \mid z_2=0)=1$, we have
$$0 = p(x_2=1\mid x_1=2) \ge p(z_2=0\mid z_1=1),$$
from which we can deduce $p(z_2=0 \mid z_1=1)=0$.

Now, we will show that $p(x_2=1\mid x_1=0)=0$, which contradicts the marginal distribution.
We have
\begin{align*}
p(x_2=1 \mid x_1=0)
&= \sum_{z_1,z_2}p(x_2=1\mid z_2)p(x_2\mid z_1)p(z_1\mid x_1=0)\\
&= p(x_2=1\mid z_2=1)p(z_2=1\mid z_1=1)p(z_1=1 \mid x_1=0),
\end{align*}
where we obtained the second equality because $p(z_1=0\mid x_1=0)=0$ and $p(z_2=0\mid z_1=1)$.
As $p(x_2=1\mid z_1=1)=0$, we have $p(x_2=1\mid z_1=1)=0\ne\frac{1}{3}$. As this is a contradiction, we have shown that there exists a marginal distribution modelable with a 3-state HMM with rank 2, but not a 2-state HMM.

\section{\label{sec:low-rank-marg}Low-Rank Hypergraph Marginalization for HMMs and PCFGs}
We provide the low-rank hypergraph marginalization algorithms for HMMs and PCFGs in
Alg.~\ref{alg:lr-marg-hmm-pcfg},
with loops over labels $z$ (and products of labels) and feature dimensions $n$ left implicit for brevity.
We also assume that the label sets for PCFG are uniform for brevity -- in practice,
this can easily be relaxed (this was not assumed in Alg.~\ref{fig:marg-hmm-pcfg}).
We show how the normalizing constants $c$ are explicitly computed using the unnormalized low-rank factors in each algorithm.

\begin{algorithm}[H]
\caption{Low-rank hypergraph marginalization for HMMs and PCFGs}
\label{alg:lr-marg-hmm-pcfg}
\begin{minipage}[t]{0.45\textwidth}
\begin{algorithmic}
\STATE {[\textit{HMM - Backward}]}
\STATE $[\tilde{V}]_{z,n} = [\phi(g(z))]_n$
\STATE $[\tilde{U}]_{z,n} = [\phi(f(z))]_n$
\STATE $[c]_z = [\tilde{U}\tilde{V}^\top\mathbf{1}]_z$
\FOR {$t \leftarrow (t+1)$ in right-to-left order}
\STATE $[\beta_{t+1}]_{z_{t+1}} = [\alpha_{t+1}]_{z_{t+1}}$
\STATE $[V_t]_{z_{t+1},n} = [\tilde{V}]_{z_{t+1},n}$
\STATE $[U_t]_{z_t,n} = p(x_t \mid z_t)[c]_{z_{t}}[\tilde{U}]_{z_t,n}$
\STATE $\alpha_t \stackrel{+}{\gets} U_t(V_t^\top\beta_{t+1})$
\ENDFOR
\STATE \textbf{return} $\alpha_0^\top \mathbf{1}$
\end{algorithmic}
\end{minipage}
\vspace{0pt}
\begin{minipage}[t]{0.50\textwidth}
\begin{algorithmic} 
\STATE {[\textit{PCFG - CKY}]}
\STATE $[\tilde{V}]_{z_1,z_2,n} = [\phi(g(z_1,z_2))]_n$
\STATE $[\tilde{U}]_{z_u,n} = [\phi(f(z_u))]_n$
\STATE $[c]_{z_u} = [UV^\top\mathbf{1}]_{z_u}$
\FOR {$(i,k) \leftarrow (i,j), (j,k)$ in span-size order}
\STATE $[\beta_{i,j,k}]_{(z_1,z_2)} = [\alpha_{i,j}]_{z_1}[\alpha_{j,k}]_{z_2}$
\STATE $[V_{i,j,k}]_{z_1,z_2,n} = [\tilde{V}]_{z_1,z_2,n}$
\STATE $[U_{i,j,k}]_{z_u,n} = [c]_{z_u}[\tilde{U}]_{z_u,n}$
\STATE $\alpha_{i,k} \stackrel{+}{\gets} U_{i,j,k}(V_{i,j,k}^\top\beta_{i,j,k})$
\ENDFOR
\STATE \textbf{return} $\alpha_{1,T}^\top \mathbf{1}$
\end{algorithmic}
\end{minipage}
\end{algorithm}

\section{Extension of the Low-Rank Constraint to Other Semirings}
Enforcing low-rank constraints in the scoring matrices $\Psi_e$
leads to a speedup for the key step in the hypergraph marginalization algorithm:
\begin{equation}
    \Psi_e\beta_v =  \left(U_e V_e^\top\right) \beta_v =  U_e\left(V_e^\top\beta_v\right),
\end{equation}
where $[\beta_v]_{z_1,z_2} = [\alpha_1]_{z_1}[\alpha_2]_{z_2}$.
While the low-rank constraint allows for speedups in both the log and probability semirings
used for marginal inference,
the low-rank constraint does not result in speedups in the tropical semiring,
used for MAP inference.
To see this, we first review the low-rank speedup in scalar form.
The key matrix-vector product step of marginal inference in scalar form is given by
\begin{align*}
\sum_{z_1,z_2} [\Psi_e]_{z_u,(z_1,z_2)}[\beta]_{z_1,z_2}
&= \sum_{z_1,z_2} \sum_n [U_e]_{z_u,n}[V_e]_{(z_1,z_2),n}[\beta]_{z_1,z_2}\\
&= \sum_n \sum_{z_1,z_2} [U_e]_{z_u,n}[V_e]_{(z_1,z_2),n}[\beta]_{z_1,z_2}\\
&= \sum_n [U_e]_{z_u,n}\sum_{z_1,z_2} [V_e]_{(z_1,z_2),n}[\beta]_{z_1,z_2},
\end{align*}
which must be computed for each $z_u$.
The first line takes $O(\mcL^{|e|+1})$ computation,
while the last line takes $O(\mcL^{|e|}N)$ computation.
The speedup comes rearranging the sum over $(z_1,z_2)$ and $n$,
then pulling out the $U_e$ factor, thanks to the distributive
propery of multiplication.
When performing MAP inference instead of marginal inference,
we take a max over $(z_1,z_2)$ instead of a sum.
Unfortunately, in the case of the max-times semiring used for MAP inference,
we cannot rearrange max and sum,
preventing low-rank models from obtaining a speedup:
\begin{align*}
\max_{z_1,z_2} [\Psi_e]_{z_u,(z_1,z_2)}[\beta]_{z_1,z_2}
&= \max_{z_1,z_2} \sum_n [U_e]_{z_u,n}[V_e]_{(z_1,z_2),n}[\beta]_{z_1,z_2}\\
&\ne \sum_n \max_{z_1,z_2} [U_e]_{z_u,n}[V_e]_{(z_1,z_2),n}[\beta]_{z_1,z_2}.
\end{align*}

\section{\label{sec:data}Data Details}
For language modeling on \textsc{Penn Treebank} (\textsc{Ptb}) \citep{ptb}
we use the preprocessing from \citet{mikolov-2011},
which lowercases all words and substitutes OOV words with UNKs. 
The dataset consists of 929k training words, 73k validation words, and 82k test words, with a vocabulary of size 10k.
Words outside of the vocabulary are mapped to the UNK token.
We insert EOS tokens after each sentence, and model each sentence, including the EOS token, independently.

The four polyphonic music datasets, Nottingham (Nott), Piano, MuseData (Muse), and JSB chorales (JSB), are used with the same splits as \citet{polyphonic}. The data is obtained via the following \href{https://github.com/pyro-ppl/pyro/blob/d7687ae0f738bd81a792dabbb18a53c0fce73765/pyro/contrib/examples/polyphonic_data_loader.py}{script}. Each timestep consists of an 88-dimensional binary vector indicating whether a particular note is played. Since multiple notes may be played at the same time, the effective vocabulary size is extremely large. The dataset lengths are given in Table~\ref{tbl:music-data}.

\begin{table}[t]
    \centering
    \begin{tabular}{lrrrr}
        \toprule
                & & \multicolumn{3}{c}{Total Length}\\
                \cmidrule{3-5}
        Dataset & Avg Len &  Train & Valid & Text \\
        \midrule
        Nott & 254.4 & 176,551 & 45,513  & 44,463 \\
        Piano& 872.5 & 75,911 & 8,540& 19,036 \\
        Muse & 467.9 &  245,202 & 82,755 & 64,339 \\
        JSB & 60.3 & 64,339 & 4,602 & 4,725\\
        \bottomrule
    \end{tabular}
    \caption{\label{tbl:music-data}The lengths for the four polyphonic music datasets. The average length of an example in the training split for each dataset is given.}
\end{table}

In experiments with PCFGs for language modeling, we also use \textsc{Ptb}, but with the splits and preprocessing used in unsupervised constituency parsing \citep{shen2018prpn,shen2018ordered,kim2019cpcfg}. This preprocessing discards punctuation, lowercases all tokens, and uses the 10k most frequent words as the vocabulary.
The splits are as follows: sections 2-21 for training, 22 for validation, 23 for test. Performance is evaluated using perplexity.

In experiments with HSMMs for video modeling, we use the \textit{primary} section of the \textsc{CrossTask} dataset \citep{zhukov2019cross}, consisting of about 2.7k instructional videos from 18 different tasks such as ``Make Banana Ice Cream'' or ``Change a Tire''. We use the preprocessing from \citet{fried2020learning}, where pretrained convolutional neural networks are first applied to extract continuous image and audio features for each frame, followed by PCA to project features to 300 dimensions.\footnote{\url{https://github.com/dpfried/action-segmentation}} We set aside 10\% of the training videos for validation.
\section{\label{sec:hsmm}Generative Process of HSMM} 

We use an HSMM to model the generative process of the sequence of continuous features for each video. The HSMM defines the following generative process: first, we sample a sequence of discrete latent states $z=(z_1, \cdots, z_K)$ with a first-order Markov model.
Next, we sample the length of observations under each state from a Poisson distribution $l_k\sim \text{Poisson}(\lambda_{z_k})$ truncated at max length $M$. The joint distribution is defined as
\begin{equation}
 \label{eqn:hsmmjoint}
     p(x, z, l) = \prod_{k=1}^K p(z_k \mid z_{k-1}) \ p(l_k \mid z_k)\prod_{i=l_1+\cdots+l_{k-1}}^{l_1+\cdots+l_k} p(x_i\mid z_k),
 \end{equation}
where the sequence length $T$ can be computed as $T=\sum_{k=1}^K l_k$. In this work, we only consider modeling continuous $x_t$, so we use a Gaussian distribution for $p(x_i\mid z_k)$.

To compute $p(x)$, we can marginalize $l, z$ using dynamic programming similar to HMMs, except that we have an additional factor of $M$: the overall complexity is $O(T\times M \times L^2)$ (ignoring the emission part since they are usually not the bottleneck). We refer to \citet{yu2010hidden} for more details.

\section{\label{sec:mlp-param}Full Parameterization of HMMs, PCFGs, and HSMMs}
In this section, we present more details on the parameterizations of the HMM, PCFG, and HSMM.
The main detail is where and how are neural networks used to parameterize state representations.

For low-rank HMMs (LHMMs) we  use the following mixed parameterization that specifically targets the state-state bottleneck:
\begin{equation}
\begin{aligned}
p(z_1 \mid z_0) &\propto \phi(f_1(\bu_{z_0}))^\top \phi(\bv_{z_1})\\
p(z_t \mid z_{t-1}) &\propto \phi(\bu_{z_{t-1}})^\top \phi(\bv_{z_t})\\
p(x_t \mid z_t) &\propto \exp(\bu_{z_t}^\top f_2(\bv_{x_t})),
\end{aligned}
\end{equation}
where $\bu_{z}$ is the embedding of $z$ when $z$ is used as head, $\bv_{z}$ its embedding when used as tail, $f_1,f_2$ are MLPs with two residual layers, and feature map $\phi(x) = \exp(Wx)$. 

The PCFG uses a similar mixed parameterization. These probabilities correspond to start ($S\to A$), preterminal ($T\to x$), and standard productions ($A\to B\ C$) respectively.
\begin{equation}
\label{eqn:pcfgparam}
\begin{aligned}
p(z_{1,N} \mid S ) &\propto \exp(f_1(\bu_{S})^\top \bu_{z_{1,N}})\\
p(x_i \mid z_i) &\propto \exp(\bu_{z_i}^\top f_2(\bv_{x_i}))\\
p( z_{i,j}, z_{j,k} \mid z_{i,k}) &\propto \begin{cases}
  \exp(\bu_{z_{i,k}}^\top \bv_{z_{i,j}, z_{j,k}}) & \substack{i+1 = j \lor \\j+1=k} \\ 
 \phi(\bu_{z_{i,k}}')^\top \phi(\bv_{z_{i,j}, z_{j,k}}) &\text{o.w.} \\
\end{cases}\\
\end{aligned}
\end{equation}
where $\bu_{z}$/$\bu_z'$ is the embedding of $z$ when $z$ is used as head,$\bv_{x}$/$\bv_{z_1, z_2}$ is the embedding of $x$/$(z_1, z_2)$ when they are used as tail, and $f_1, f_2$ are MLPs with two residual layers as in \citet{kim2019cpcfg}. We use the feature map $\phi(x) = \exp(Wx - \|x\|_2^2/2)$.

For both HMMs and neural PCFG models, we use the same parameterization of the MLPs $f_1$ and $f_2$ as \citet{kim2019cpcfg}:
\begin{equation}
\label{eqn:res}
\begin{aligned}
f_i(x) &= g_{i,1}(g_{i,2}(W_i x)),\\
g_{i,j}(y) &= \ReLU(U_{i,j} \ReLU(V_{i,j} y)) + y,
\end{aligned}
\end{equation}
with $i,j \in \set{1,2}$, and
$W_i,V_{i,j},U_{i,j} \in \R^{D \times D}$.

For HSMMs, the baseline ({HSMM} in Table~\ref{tbl:cky}) follows the fully unsupervised setting in \citet{fried2020learning} except that we don't apply any state constraints from the prior knowledge of each task.\footnote{We got rid of those constraints to allow for changing the total number of states, since otherwise we can't make any changes under a predefined state space.} The model maintains a log transition probability lookup table for $p(z_k \mid z_{k-1})$, a lookup table for the log of the parameters of the Poisson distribution $\lambda_{z_k}$. We maintain a mean and a diagonal covariance matrix for the Gaussian distribution $p(x_i\mid z_k)$ for each $z_k$. For low-rank HSMMs (LHSMMs), we use the same parameterization for $p(z_k \mid z_{k-1})$ as in HMMs:
\begin{equation}
    p(z_k \mid z_{k-1}) \propto \phi(\bu_{z_{t-1}})^\top \phi(\bv_{z_t}),
\end{equation}
where $\bu_{z}$ is the embedding of $z$ when $z$ is used as head, $\bv_{z}$ its embedding when used as tail, and the feature map $\phi(x) = \exp(Wx)$. The emission parameterization is the same as in baseline HSMMs, a Gaussian kernel.

\section{\label{sec:opt}Initialization and Optimization Hyperparameters}
We initialize the parameters $W$, in $\phi(x) = \exp(Wx)$ and variants, of feature maps using orthogonal feature projections \citep{choromanski2020performer}, and update it alongside the model parameters during training.

HMM parameters are initialized with the Xavier initialization \citep{glorot2010understanding}.\footnote{
For banded experiments, we initialize the band parameters by additionally adding 30 to each element.
Without this the band scores were too small compared to the exponentiated scores, and were ignored by the model.
}
We use the AdamW \citep{adamw} optimizer with a learning rate of $0.001$, $\beta_1=0.9,\beta_2=0.999$, weight decay 0.01, and a max grad norm of 5.
We use a state dropout rate of 0.1, and additionally have a dropout rate of 0.1 on the feature space of LHMMs.
We train for 30 epochs with a max batch size of 256 tokens, and anneal the learning rate by dividing by 4 if the validation perplexity fails to improve after 4 evaluations. Evaluations are performed 4 times per epoch.
The sentences, which we model independently from one another, are shuffled after every epoch.
Batches of sentences are drawn from buckets containing sentences of similar lengths to minimize padding. 

For the polyphonic music datasets, we use the same hyperparameters as the language modeling experiments, except a state dropout rate of 0.5 for JSB and Nottingham, 0.1 for Muse and Piano. We did not use feature space dropout in the LHMMs on the music datasets. For Nottingham and JSB, sentences were batched in length buckets, the same as language modeling. Due to memory constraints, Muse and Piano were processed using BPTT with a batch size of 8 for Muse and 2 for Piano, and a BPTT length of 128. We use $D=256$ for all embeddings and MLPs on all datasets, except Piano, which due to its small size required $D=64$ dimensional embeddings and MLPs. 

For PCFGs, parameters are initialized with the Xavier uniform initialization \citep{glorot2010understanding}.
We follow the experiment setting in \citet{kim2019cpcfg} and use the Adam \citep{kingma2017adam} optimizer with $\beta_1=0.75,\beta_2=0.999$, a max grad norm of 3, and we tune the learning rate from $\{0.001, 0.002\}$ using validation perplexity. We train for 15 epochs with a batch size of 4. The learning rate is not annealed over training, but a curriculum learning approach is applied where only sentences of at most length 30 are considered in the first epoch. In each of the following epochs, the longest length of sentences considered is increased by 1.

For HSMMs, we use the same initialization and optimization hyperparameters as \citet{fried2020learning}: The Gaussian means and covariances are initialized with empirical means and covariances (the Gaussian parameters for all states are initialized the same way and they only diverge through training). The transition matrix is initialized to be uniform distribution for baseline HSMMs, and the transition embeddings are initialized using the Xavier initialization for LHSMMs. The log of Poisson parameters are initialized to be 0. We train all models for 4 epochs using the Adam optimizer with initial learning rate of 5e-3, and we reduce the learning rate 80\% when log likelihood doesn't improve over the previous epoch. We clamp the learning rate to be at least 1e-4. We use a batch size of 5 following \citet{fried2020learning}, simulated by accumulating gradients under batch size 1 in order to scale up the number of states as much as we can. Gradient norms are clipped to be at most 10 before updating. Training take 1-2 days depending on the number of states and whether a low-rank constraint is used.

We use the following hardware for our experiments: for HMMs we run experiments on 8 Titan RTX GPUs with 24G of memory on an internal cluster. For PCFGs and HSMMs we run experiments on 1 Nvidia V100 GPU with 32G of memory on an internal cluster.

\section{HMM Rank Analysis}
\label{sec:rank}

Table~\ref{tbl:hmm-rank} contains the empirical ranks of trained HMMs and LHMMs, estimated by counting the number of singular values greater than 1e-5. Note that the feature dimension $N$ is the maximum attainable rank for the transition matrix of an LHMM. Although LHMMs often manage to achieve the same validation perplexity as HMMs at relatively small $N$, the ranks of the transition matrices are much lower than both their HMM counterparts as well as $N$.
At larger state sizes, the ranks of learned matrices are almost half of their max achievable rank. Interestingly, this holds true for HMMs as well, with the empirical rank of the transition matrices significantly smaller than the number of states. Whether this implies that the models can be improved is left to future investigations.

\begin{table}[t]
    \centering
    \begin{tabular}{lllrrr}
    \toprule
        Model & $L$ & $N$ & rank$(A)$ & rank$(O)$ & Val PPL  \\
        \midrule
        HMM & 16384 & - & 9187 & 9107 & 144\\ 
        LHMM & 16384 & 8192 & 2572 & 7487 & 141\\
        LHMM & 16384 & 4096 & 2016 & 7139 & 144\\
        LHMM & 16384 & 2048 & 1559 & 6509 & 141\\
        \midrule
        LMM & 8192 & - &  5330 & 5349 & 152\\
        LHMM & 8192 & 4096 & 1604 & 5113 & 149\\
        LHMM & 8192 & 2048 & 1020 & 4980 & 153\\
        LHMM & 8192 & 1024 & 791 & 5033 & 161\\
		\midrule
        HMM & 4096 & - & 2992 & 3388 & 155\\
        LHMM & 4096 & 2048 & 1171 & 3300 & 154\\
        LHMM & 4096 & 1024 & 790 & 2940 & 156\\
        LHMM & 4096 & 512 & 507 & 3186 & 163\\
        \bottomrule
    \end{tabular}
    \vspace{0.5em}
    \caption{\label{tbl:hmm-rank}Ranks and validation perplexities for HMMs and LHMMs. The number of states is given by $L$ and the dimensionality of the feature space by $N$. The HMM uses softmax for the emission, and therefore does not have a value for $N$. The transition matrix is denoted by $A$, and the emission matrix by $O$. The rank was estimated by counting the number of singular values greater than 1e-5.
    Models were trained with 0.1 state and feature dropout.}
\end{table}

\section{Low-rank and Banded HMM Parameterization}
\label{sec:banded}
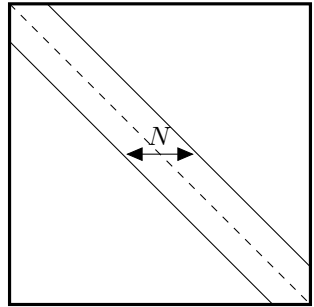
\begin{figure}
    \centering
    \begin{tikzpicture}
    \draw[very thick] (2,2)--(-2,2)--(-2,-2)--(2,-2)--cycle;
    \draw[dashed](-2,2)--(2,-2);
    \draw (-1.5,2)--(2,-1.5);
    \draw (-2,1.5)--(1.5,-2);

    \draw[<->] (-.45,0)--node[above]{$N$}(.45,0);
    \end{tikzpicture}
    \caption{An example of a banded matrix with width $N$, which has $N/2$ nonzero elements on both sides of the diagonal for each row.}
    \label{fig:banded}
\end{figure}

In some scenarios, the low-rank constraint may be too strong. For example, a low-rank model is unable to fit the identity matrix, which would have rank $L$. 
In order to overcome this limitation, we extend the low-rank model while preserving the computational complexity of inference.
We perform experiments with an additional set of parameters $\theta \in\R^{L\times L}$ which allow the model to learn high-rank structure (the experimental results can be found in Tbl.~\ref{tbl:hmm}).
We constrain $\theta$ to have banded structure,
such that $[\theta]_{z_{t-1},z_t} = 0$ if $|z_t - z_{t-1}| > N/2$.
See Fig.~\ref{fig:banded} for an illustration of banded structure.

Let band segment $B_{z} = \set{z' : |z - z'| \le N/2}$.
The transition probabilities are then given by
\begin{equation}
    p(z_t \mid z_{t-1}) = \frac{[\theta]_{z_{t-1},z_{t}} + \phi(\bu_{z_{t-1}})^\top \phi(\bv_{z_{t}})}{Z_{z_{t-1}}},
\end{equation}
with normalizing constants
\begin{equation}
\begin{aligned}
Z_{z_{t-1}}
&= \sum_{z_{t}} [\theta]_{z_{t-1},z_{t}} + \phi(\bu_{z_{t-1}})^\top \phi(\bv_{z_{t}})\\
&= \sum_{z_{t}\in B_{z_{t-1}}} [\theta]_{z_{t-1},z_{t}} + 
    \phi(\bu_{z_{t-1}})^\top \sum_{z_t} \phi(\bv_{z_{t}}).
\end{aligned}
\end{equation}
The normalization constant for each starting state $Z_{z_{t-1}}$ can be computed in time $O(N)$.

This allows us to perform inference quickly.
We can use the above to rewrite the score matrix $\Psi_t \propto \theta + UV^\top$, which turns the inner loop of Eqn.~\ref{eqn:hypergraph-update-kernel-matrix} (specialized to HMMs) into
\begin{equation}
    \label{eqn:hmm-band-update}
    \alpha_t = \Psi_t\beta_{t+1} \propto (\theta + UV^\top)\beta_{t+1} = \theta\beta_{t+1} + U(V^\top \beta_{t+1}),
\end{equation}
omitting constants (i.e. emission probabilities and normalizing constants).
Since $\theta$ is banded, the banded matrix-vector product $\theta\beta_t$ takes time $O(LN)$.
This update, in combination with the low-rank product, takes $O(LN)$ time total. Each update in the hypergraph marginalization algorithm is now 3 matrix-vector products costing $O(LN)$ each, preserving the runtime of inference.

\section{Music Results}
\label{sec:full-music}
The full results on the polyphonic music modeling task can be found in Tbl.~\ref{tbl:full-music}, with additional models for comparison.
Aside from the RNN-NADE~\citep{polyphonic}, which models the full joint distribution of notes as well as temporal dependencies;  autoregressive neural R-Transformer~\citep{rtransformer} (as reported by \citet{betalstm}) and LSTM~(as reported by \citet{flow}); latent continuous LV-RNN \citep{nasmc} and SRNN \citep{srnn}; and latent discrete TSBN \citep{tsbn} and the baseline HMM; we additionally include the autoregressive Seq-U-Net \citet{sequnet}, the continuous latent STORN \citep{storn}, DMM \citep{dmm} and LNF \citep{flow}.

\begin{table}[t]
\centering
\begin{tabular}{lcccc}
\toprule
Model       & Nott & Piano & Muse & JSB \\
\midrule
RNN-NADE & 2.31  & 7.05        & \textbf{5.6}        & 5.19          \\
Seq-U-Net & 2.97 & \textbf{1.93} & 6.96 & 8.173 \\
R-Transformer & \textbf{2.24} & 7.44 & 7.00 & 8.26 \\
LSTM  & 3.43 & 7.77   & 7.23 & 8.17     \\
STORN    & 2.85       & 7.13        &  6.16       & 6.91  \\
LV-RNN    & 2.72       & 7.61        & 6.89       &\textbf{ 3.99}\\
SRNN     & 2.94       & 8.2         & 6.28       & 4.74          \\
DMM     & 2.77 &    7.83 &  6.83 &  6.39 \\
LNF  & 2.39  & 8.19 & 6.92  & 6.53    \\
\midrule
TSBN     & 3.67       & 7.89        & 6.81       & 7.48          \\
HMM &  2.43 & 8.51 & 7.34 & 5.74 \\
LHMM & 2.60 & 8.89 & 7.60 & 5.80 \\
\bottomrule
\end{tabular}
\vspace{0.5em}
\caption{\label{tbl:full-music}
Polyphonic music negative log-likelihood, measured in nats.
The HMM models have $\mcL=2^{11}$ states and the LHMM has rank $N=2^{9}$, a 4:1 state:rank ratio.
}
\end{table}

\begin{figure*}[!tp]
  \centering
  \begin{subfigure}[t]{0.45\textwidth}
  \centering
  \includegraphics[width=0.9\textwidth]{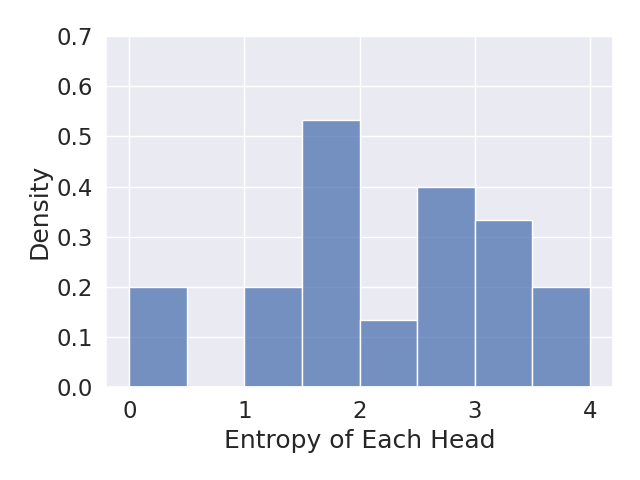}
  \caption{Softmax Parameterization}
  \end{subfigure}
  \begin{subfigure}[t]{0.45\textwidth}
  \centering
  \includegraphics[width=0.9\textwidth]{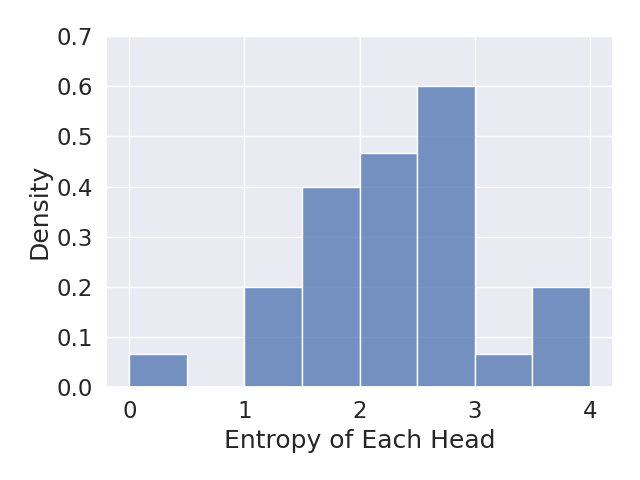}
  \caption{Low-rank Parameterization}
  \end{subfigure}
  \caption{\label{fig:example_production}Histogram of entropies of $P(B\ C\mid A)$. The average entropy is 2.26 for softmax and 2.34 for the low-rank parameterization. We use $|\mcN|=30$, $|\mcP|=60$, and $N=16$ for the rank.}
\end{figure*}
\section{PCFG Analysis}
\begin{table}[!htp]
\centering
\begin{tabular}{@{}ccccr@{}}
\toprule
\multicolumn{4}{c}{Kernel for $B\ C $} & \multirow{2}{*}{PPL}\\
\cmidrule{1-4}
$\mcN\times \mcN$ & $\mcN\times\mcP$  & $\mcP\times\mcN$ & $\mcP\times\mcP$ \\
\midrule
  SM & SM & SM & SM & 243.19\\
  LR & SM & SM & SM & 242.72\\
  LR & LR & LR & SM &  259.05 \\
  LR & LR & LR & LR & 278.60 \\
\bottomrule
\end{tabular}
\vspace{0.5em}
\caption{\label{tbl:cky-ablation}Model perplexities evaluated on the validation set of \textsc{Ptb}. Here we use $|\mcN|=30$, $|\mcP|=60$, and $N=16$ rank. SM denotes the use of softmax, while LR a low-rank factorization.}
\end{table}

Figure~\ref{fig:example_production} shows the entropy distribution of the production rules $H(P(B\ C | A))$ for both using softmax kernel and the approximation. The average entropies of the two distributions are close. Besides, under this setting, $P(B\ C \in \mcN \times \mcN| A)$ are close for both kernels as well (softmax 0.20, linear 0.21), eliminating the possibility that the kernel model simply learns to avoid using $B\ C \in \mcN \times \mcN$ (such as by using a right-branching tree).

In Table~\ref{tbl:cky-ablation}, we consider the effects of the mixed parameterization, i.e. of replacing the softmax parameterization with a low-rank parameterization. In particular, we consider different combinations of preterminal / nonterminal tails $B\ C\in \mcN\times\mcN$, $B\ C\in \mcN\times\mcP$, $B\ C\in \mcP\times\mcN$, and $B\ C\in \mcP\times\mcP$ (our main model only factorizes nonterminal / nonterminal tails). Table~\ref{tbl:cky-ablation} shows that we get the best perplexity when we only use $K$ on $B\ C\in \mcN\times\mcN$, and use softmax kernel $K_\SM$ for the rest of the space. This fits with previous observations that when the label space $|\mcL|$ is large, a model with a very small rank constraint hurts performance.\footnote{In this particular ablation study, the size of $\mcN\times\mcN$ is only one-ninth of the total state space size $\{\mcN \cup \mcP\}\times \{\mcN \cup \mcP\}$.}

\section{\label{sec:frontier} Speed and Accuracy Frontier Analysis}
\begin{figure}[H]
\centering
\begin{subfigure}[t]{0.48\textwidth}
\centering
\includegraphics[height=4.5cm]{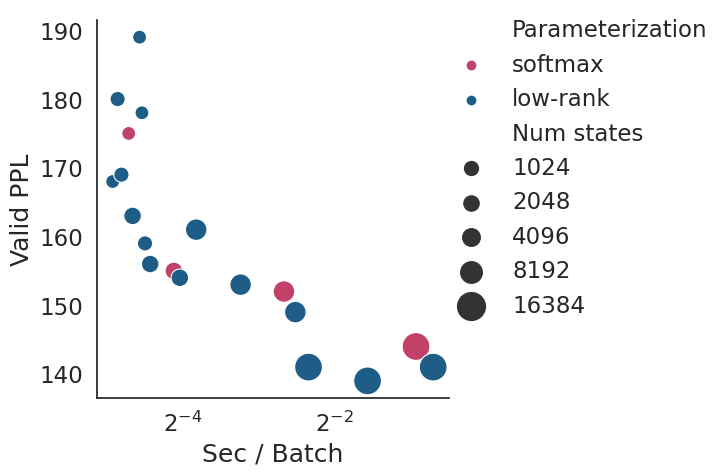}
\end{subfigure}
\hspace{1em}
\begin{subfigure}[t]{0.48\textwidth}
\centering
\includegraphics[height=4.5cm]{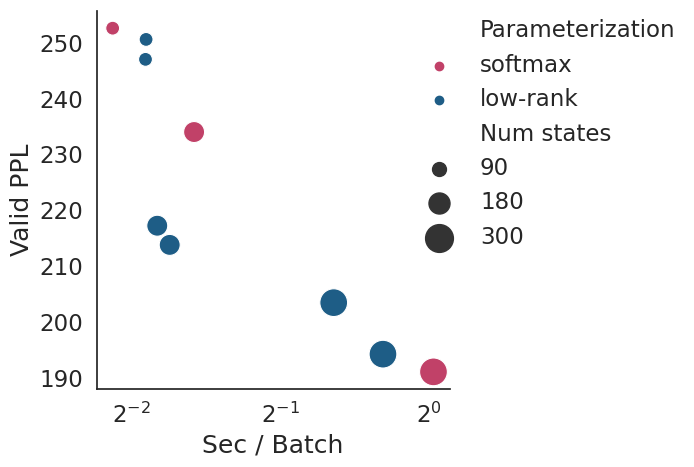}
\end{subfigure}
\caption{
\label{fig:frontier}
The speed, in seconds per batch, versus accuracy, in perplexity,
for HMMs (left), PCFGs (right), and low-rank versions over a range of model sizes.
As lower is better for both measures of speed and accuracy, the frontier is the bottom left.
}
\end{figure}

We provide plots of the speed and accuracy over a range of model sizes for HMMs and PCFGs,
in Figure~\ref{fig:frontier} (left and right respectively).
Speed is measured in seconds per batch, and accuracy by perplexity.
Lower is better for both.

For HMMs, we range over the number of labels $L\in\set{2^{10},2^{11},2^{12},2^{13},2^{14}}$.
For softmax HMMs, more accurate models are slower, as shown in Figure \ref{fig:frontier} (left).
However, we find that for any given accuracy for a softmax model,
there exists a similarly accurate LHMM that outspeeds it.
While we saw earlier in Figure \ref{fig:hmm-ppl-features-dropout}
that at smaller sizes
the low-rank constrain hurt accuracy, a model with a larger state size but
lower rank achieves similar accuracy at better speed compared to a small HMM.

For PCFGs, we range over $L\in\set{90,180,300}$.
We find a similar trend compared to HMMs: accuracy results in slower models,
as shown in Figure \ref{fig:frontier} (right).
However, the LPCFG does not dominate the frontier as it did with HMMs.
We hypothesize that this is because of the small number of labels in the model.
In the case of HMMs, smaller softmax HMMs were more accurate than the faster low-rank versions,
but larger LHMMs with low rank were able to achieve similar perplexity at faster speeds.
This may be realized by exploring LPCFGs with more state sizes,
or simply by scaling further.

\section{Potential Negative Impact}
\label{sec:impact}
While work on interpretable and controllable models is a step towards machine that can more easily be understood by and interact with humans, introducing external-facing components leaves models possibly more vulnerable to adversarial attacks. In particular, the interpretations (in conjunction with the predictions) afforded by interpretable models may be attacked \citep{adv-interp}. Additionally, models with simple dependencies may be easier for adversaries to understand and then craft attacks for \citep{Zhang2021LabelFA,adv-phmm}.
\end{document}